\documentclass[acmsmall,screen]{acmart}
\acmJournal{CSUR}
\usepackage{float}
\usepackage{times}  
\usepackage{helvet} 
\usepackage{courier}  

\usepackage{booktabs} 
\usepackage{multirow}
\usepackage{subfigure}
\usepackage{mathrsfs}
\usepackage[ruled,vlined]{algorithm2e}
\usepackage{algorithm2e}
\usepackage{balance} 
\usepackage{color}
\usepackage{enumitem}
\usepackage{textcomp}
\usepackage{xcolor}
\usepackage{makecell}
\usepackage{bm}
\usepackage{balance}
\usepackage{tabu}
\usepackage{amsthm,amsmath}
\usepackage{mathrsfs}
\usepackage{hhline}
\usepackage{epstopdf}
\usepackage{natbib}

\AtBeginDocument{%
  }

\begin{document}

\title{Deep Reinforcement Learning for Demand Driven Services in Logistics and Transportation Systems: A Survey}


\author{Zefang Zong}
\author{Jingwei Wang}
\affiliation{%
  \institution{Department of Electronic Engineering, BNRist, Tsinghua University}
  \country{China}
}
\email{zongzf19@mails.tsinghua.edu.cn}

\author{Tao Feng}
\affiliation{%
  \institution{University of Illinois at Urbana-Champaign}
  \country{USA}
}
\email{taofeng2@illinois.edu}

\author{Tong Xia}
\affiliation{%
  \institution{Department of Computer Science, the University of Cambridge}
  \country{UK}
}

\author{Yong Li}
\affiliation{%
  \institution{Department of Electronic Engineering, BNRist, Tsinghua University}
  \country{China}
}
\email{liyong07@tsinghua.edu.cn}

\newcommand{\zong}[1]{{\color{black} #1}}

\begin{abstract}
  Recent technology development brings the boom of numerous new Demand-Driven Services (DDS) into urban lives, including ridesharing, on-demand delivery, express systems and warehousing. In DDS, a service loop is an elemental structure, including its service worker, the service providers and corresponding service targets. The service workers should transport either people or parcels from the providers to the target locations. 
Various planning tasks within DDS can thus be classified into two individual stages: 1) Dispatching, which is to form service loops from demand/supply distributions, and 2) Routing, which is to decide specific serving orders within the constructed loops. Generating high-quality strategies in both stages is important to develop DDS but faces several challenges. Meanwhile, deep reinforcement learning (DRL) has been developed rapidly in recent years. It is a powerful tool to solve these problems since DRL can learn a parametric model without relying on too many problem-based assumptions and optimize long-term effects by learning sequential decisions.   
In this survey, we first define DDS, then highlight common applications and important decision/control problems within. For each problem, we comprehensively introduce the existing DRL solutions.
We also introduce open simulation environments for development and evaluation of DDS applications. Finally,  we analyze remaining challenges and  discuss further research opportunities in DRL solutions for DDS. 
\end{abstract}




\maketitle
\section{Introduction}

 The continuous urbanization and development of mobile communication have brought many new application demands into urban daily lives.  Among all, the services that transport either humans or parcels to provided destinations following individual demands or system requirements are critical in both urban logistics and transportation nowadays. We define such kinds of services as Demand-Driven Services (DDS). 
  For example, the on-demand food delivery service as a typical DDS is widely used since it improves dietary convenience significantly. More than 30 million orders are generated every day on the Meituan-Dianping platform, one of the world’s largest on-demand delivery service providers~\cite{meituan}. As another example, large-scale online ridesharing services such as Uber and DiDi have substantially transformed the transportation landscape, offering huge opportunities for boosting the current transportation efficiency. These DDS applications provide striking efficiency to city operations in both logistics and transportation as well as many opportunities to related research fields. Intelligent control with minimal manual intervention upon DDS systems is critical to guarantee their effectiveness and has drawn many research interests.

\begin{figure}[ht]
    \vspace{-2mm}
    \centering
    \includegraphics[width = 0.60\textwidth]{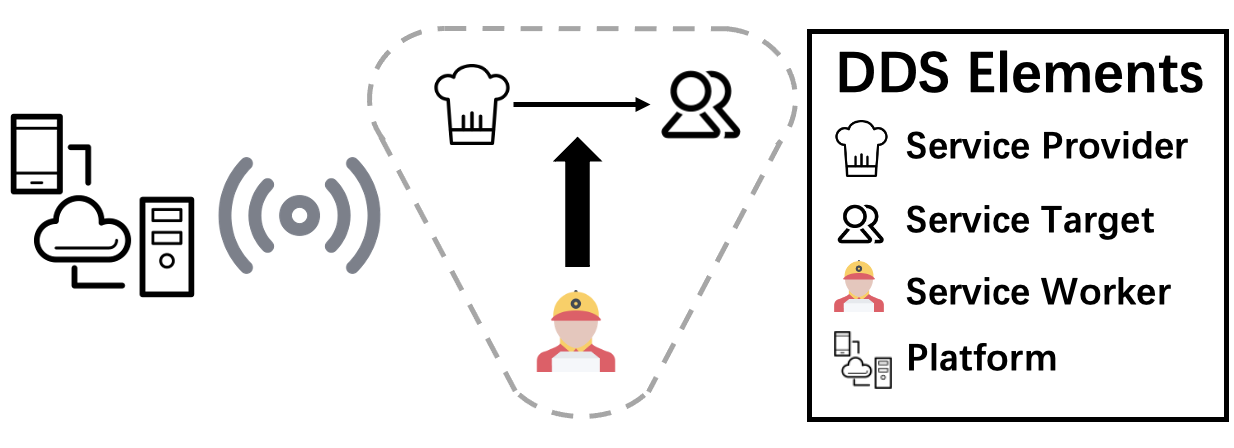}
    \vspace{-2mm}
    \caption{The visualization of two independent service loops using instant delivery as an example. The restaurant, customer, and courier serve as the service provider, service target, and service worker. 
    }
    
    \vspace{-4mm}
    \label{fig:loop}
\end{figure}

 In a typical DDS task, there are several roles involved that an implemented system should consider, including the service workers, service providers, and corresponding targets. For example, in an on-demand delivery system, a man who orders food can be seen as a service target, while the restaurant from which the food is ordered is the service provider. A group of such delivery tasks are then assigned to and accomplished by a courier, i.e., the worker. These core DDS elements form a DDS loop, and such an example is illustrated in Figure~\ref{fig:loop}. The same formulation can also be constructed within the ride-sharing scenario. Each customer who calls for a ride has his/her destination as the target, and the driver serves as the service worker. The DDS platforms that support either delivery or ride-sharing services are supposed to provide corresponding algorithms to 1) construct reasonable service loops and 2) guide workers to complete assignments within loops. We show DDS loop formulations in several typical scenarios in Table~\ref{tab:loops}, including on-demand delivery, ridesharing, express systems, and warehousing.

\begin{table}[t]
    \centering
    \caption{Elements in a DDS loop of several typical DDS scenarios. AGV is the abbreviation for Autonomous Guiding Vehicle. 
    \vspace{-2mm}
    }
    \resizebox{0.75\textwidth}{!}
    {
    \begin{tabular}{c||c|c|c}
        \Xhline{1pt} 
        \multirow{2}{*}{DDS Scenario}&\multicolumn{3}{c}{DDS Loop}\\
        \cline{2-4}
         &Service Provider&Service Target&Service Worker\\
         \hline 
         On-demand Delivery &Restaurant & Customer & Courier\\
         Ridesharing &Passenger Origin & Destination & Driver\\
         Express (Sending) & Consignor &Depot&Courier\\
         Express (Delivery) & Depot & Consignee&Courier\\
         Warehousing &Shelf, Entry, Station &Shelf, Entry, Station &AGV\\
         \Xhline{1pt} 
    \end{tabular}
    }

    \label{tab:loops}
\end{table}

With fundamental DDS elements defined, how to manage different service demand pairs (providers and targets), schedule available service workers and control the entire service system become the major objectives of developing a centralized intelligent DDS platform. Major research problems can be classified into two aspects. First, forming DDS loops upon demand pairs and workers, which is also named \textit{Dispatching}
, is the first-hand challenge to deal with. 
The loop forming process, i.e., the \textit{Matching} between demands and workers can be originated from the traditional bipartite graph matching problem, while the dynamic features in the entire environment bring much more complexity. A good dispatching mechanism should not only consider the current states of workers with scattered demands but also take future distributions into account for long-term optimization.
Furthermore, even a worker is not matched with service demands at present, there still remains a large action space to arrange the idle workers into other areas, which forms the \textit{Fleet Management} problem. The loop-forming stage can be seen as the first stage for the complete DDS.

\begin{figure*}[ht]
    \centering
    \vspace{-4mm}
    \includegraphics[width=0.75\textwidth]{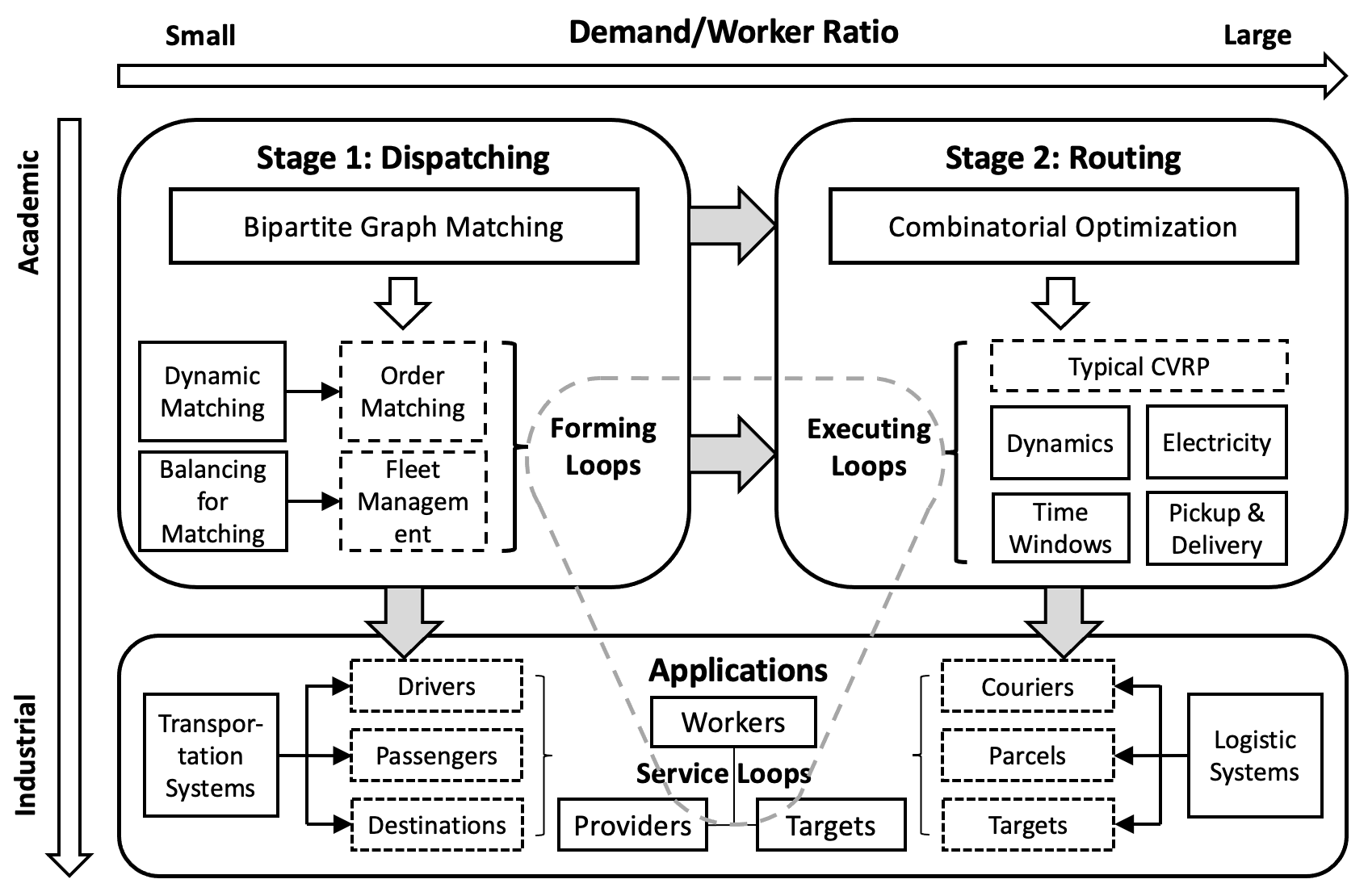}
    \vspace{-4mm}
    \caption{The overview of DDS problems, including the dispatching stage and the routing stage. We demonstrate the transformation from originated mathematical formulation to industrial applications in the vertical axis, and distinguish the two different planning stages in the horizontal axis. Note that the two stages are not rigidly separated, but such a classification is necessary to concentrate on primary challenges in different practical scenarios. A low demand/worker ratio implies that the primary challenge is to determine how workers and demands should be matched, while a large one indicates that the major optimization space lies in the routing stage. We will discuss such a relationship in details in Sec~\ref{Sec_background}.3. \zong{CVRP here refers to the capacitated vehicle routing problems with four common variants below, which will be introduced in detail in Sec 4.}
    }
    \label{fig:roadmap}
\end{figure*}

\zong{Second, after being assigned with numerous demands to satisfy, how to execute formed loops, i.e., to plan and optimize the visiting orders of the demand set as \textit{Routing}, is also critical to determine the entire system efficiency. }The routing problem can be originated from the conventional Traveling Salesman Problem ~\cite{flood1956traveling}, where a salesman is supposed to visit all cities without revisiting anyone of them. The further Vehicle Routing Problems (VRP)~\cite{dantzig1959truck}, and its variants are valuable in the mathematical formulation of most real-world routing scenarios~\cite{schneider2014electric,desrosiers1984routing,psaraftis1988dynamic,min1989multiple}. A high-quality routing strategy should minimize the total traveling distance to decrease the expenses of the workers. The routing stage can be seen as the second stage after dispatching. 
A robust and stable routing strategy generation is also important to provide decision information back to the dispatching stage. 
We illustrate the relationship between the two stages in Figure~\ref{fig:roadmap}.

The solutions to the mathematical formulations
of both two stages were widely studied previously. For instance, the Kuhn-Munkres (KM) algorithm for Bipartite-Graph Matching and Branch-and-Bound for TSP and VRP could provide exact solutions for simple static problems with limited scales~\cite{munkres1957algorithms, toth2002branch}. Considering multiple real-world constraints and additional factors, more complicated dispatching and routing problems are also further investigated extensively in the field of operations research, applied maths, etc~\cite{liao2003real,ozkan2020dynamic,macs-colony:a,recreatesearch}. In complicated scenarios with larger problem scales, exact optimizations are almost impossible to obtain. Meanwhile, heuristics and meta-heuristics were widely accepted as an alternative to generate approximate solutions within a much more reasonable time in both stages of DDS~\cite{hu2021dynamic, favaretto2007ant}. These heuristics-based methods could generate satisfactory solutions in online scenarios and are thus practical in many real-world DDS systems. However, there is still much potential in exploring better solutions with higher quality, higher efficiency on larger scales.

As machine learning shows astonishing performance in recent years, it is of great potential to utilize the learning-based techniques to further develop DDS systems. Some researchers attempted to train generation models to learn the mapping between problem distributions and their optimal solutions directly~\cite{nowak2017note, xing2020graph,joshi2019efficient}. The optimal solutions are required as the labels, which are usually given as demonstrations from other expert algorithms. Such a supervised learning mechanism could achieve significant performances cooperated with proper neural network designs, such as Graph Neural Networks (GNN). However, providing expert demonstrations as training signals directly have natural limitations on the trained model. The performance of the learnt solution is bounded by the expert, and the learnt policy would favor only one solution even if multiple solutions are equally good. Meanwhile, Reinforcement Learning (RL) methods learn policies from experience rather than demonstrations, and have been applied in many planning tasks~\cite{sutton2018reinforcement}. RL could generate strategies by modeling a decision process as a Markov Decision Process (MDP).  A predefined reward from a long-term perspective works as the feedback signal to any action attempts so that RL can optimize sequential decisions.  The trial-and-error process could train the agent to learn to select the best action corresponding to different inner states and outside environments. As deep neural networks provides much stronger ability on feature representative and pattern recognition, combining neural networks and RL shows great performances~\cite{mnih2015human}. Many deep RL (DRL) algorithms are further proposed and become state-of-the-art frameworks in control and scheduling tasks. DRL does not have to rely on manually designed assumptions and features by training a parameterized model to learn the optimal control.  It is trivial to consider using it as the structure for solving the series of DDS tasks. 

\begin{table}[t]
    \centering
    \caption{Comparison of reinforcement learning against supervised learning in solving DDS problems.}
    \begin{tabular}{c|ccc}
    \Xhline{1pt} 
        Learning Mechanism& Learning Signal & Performance Bound & Solution Flexibility \\
    \hline
        Supervised Learning & Demonstration& Expert & Single Solution\\
        Reinforcement Learning & Experience& Potential over Expert& Multiple Solutions \\
    \Xhline{1pt} 
    \end{tabular}
    
    \label{tab:RLSL}
\end{table}

\begin{figure}[ht]
    \centering
    \includegraphics[width=0.80\textwidth]{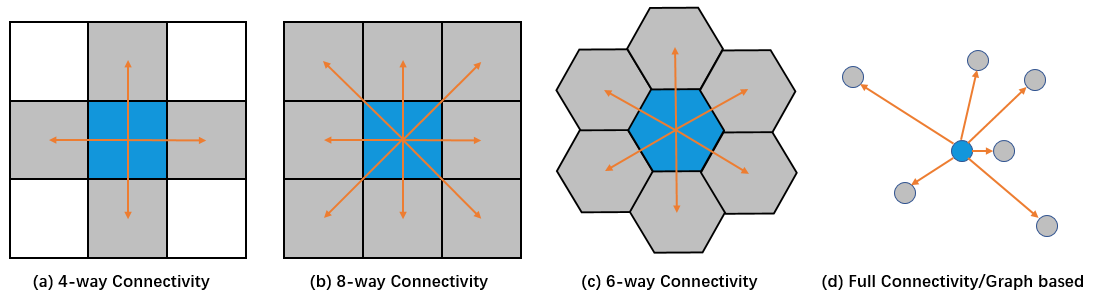}
    \caption{A sample of different grid-based navigation and partitioning schemes: (a) 4-way connectivity through cardinal directions, (b) 8-way connectivity with ordinal directions, (c) 6-way connectivity using hexagon-based representations. (d) Full connectivity, can also be modeled as a graph structure. }
    \label{fig:grid}
\end{figure}

In this survey, we focus on how DRL can benefit to the development of DDS systems in both the dispatching stage and the routing stage respectively.
We first introduce major DRL algorithms and four typical DDS systems in urban operations. Then we summarize existing DRL based solutions according to the following dimensions: 
\begin{itemize}
    \item \textbf{Problem.} We classify the research problems in both dispatching and routing stages into more precise sub-problems. \textit{Order dispatching} along with \textit{fleet management} included in the dispatching stage are investigated. As for the routing stage, we first introduce the ones solving typical Capacitated VRP (CVRP) as mathematical solutions, while more practical solutions for VRP variants are also discussed. We consider four variant problems with additional constraints in this survey, including dynamic VRP (DVRP), electric VRP (EVRP), VRP with Time Windows (VRPTW) and VRP with pickup and delivery (VRPPD).
    \item \textbf{Scenario.} The aforementioned research problems exist in several applicable scenarios, and four common DDS scenarios are included in this survey. In transportation systems, we introduce ridesharing services, where vehicles are assigned to transport passengers to their destinations. Specifically, ridesharing can be further classified into ride-hailing, where each driver serves only one passenger in a loop, and ride-pooling, where multiple passengers can share a ride at the same time. As for the logistic systems where parcels are transported from providers to targets, we summarize solutions for both on-demand delivery systems that fulfill people's instant demands and traditional express systems with longer service duration. We also introduce modern warehousing systems where Autonomous Guiding Vehicles (AGVs) transport parcels from locations to another. Note that some important literature providing solutions within a mathematical formulation is also included~\cite{NIPS18-VRP,ICLR19}. 
    \item \textbf{Algorithm.} We distinguish the detailed RL algorithm used during model training. Most commonly used ones in existing works belong to model-free RL methods, including DQN~\cite{mnih2015human}, PPO~\cite{schulman2017proximal}, REINFORCE~\cite{williams1992simple}, etc. We also discuss whether the DDS task is constructed as a single agent MDP or a multi-agent one.
    \item \textbf{Network Structure.} We also distinguish the neural network design in each literature. Commonly used networks include Convolutional Neural Networks (CNN), Graph Neural Networks (GNN) and its variants (including GCN and others), attention (ATT) based networks and its variants (including single$\verb|/|$ multi head attentions).
    \item \textbf{Data Type and Data Scheme.} We indicate the data type used in each literature, either real-world or generated based on pre-defined random seeds with a given distribution. Meanwhile, spatial locations of data are utilized in several ways for simplification to a different extent.  There are four data schemes originated from the real road networks as follows: 4-way connectivity with cardinal directions, 8-way connectivity with ordinal directions, 6-way connectivity based on hexagon-grids, and original discrete graph-based structure. The first two can also be summarized as square-grids.  Different data schemes are shown in Figure~\ref{fig:grid}. 
    \item \textbf{Data and Code Availability.} To present the extent of reproducibility of the investigated literature, we report the data availability of the proposed methods. A checkmark means that the data is released by the researchers or could be easily found via a direct web search. We also report the availability of the code. Both original open-sourced codes and re-implementation from the third party are considered.
\end{itemize}
Besides, we also introduce the available simulation environments for DDS, which is critical to simulate real-world scenarios with much fewer expenses. Finally, several challenges of using DRL to solve DDS and remaining open research problems are summarized. 

Previous literature investigating relative applications via DRL includes surveys by Haydari et al.~\cite{haydari2020deep}, Qin et al.~\cite{qin2021reinforcement}, Yan et al.~\cite{yan2022reinforcement} and several reviews on VRP~\cite{mazyavkina2021reinforcement}, as shown in Table~\ref{tab:surveys}. However, Haydari et al.~\cite{haydari2020deep} focused on the general planning problems in Intelligent Transportation Systems from where Transportation Signal Control (TSC) and Autonomous Driving are emphasized. \zong{Qin et al.~\cite{qin2021reinforcement} only investigated the dispatching problems in ridesharing scenarios. Conversely, Yan et al.~\cite{yan2022reinforcement} only dives into the logistics and supply chain management applications without discussing other services where people are transported such as dispatching.} Mazyavkina et al.~\cite{mazyavkina2021reinforcement} introduced DRL solutions on mathematical VRPs included in more general combinatorial optimizations. In contrast, we are the first to define DDS from a practical system level and classify specific research problems in several scenarios with DRL-based solutions. Meanwhile, compared to existing reviews on general DRL algorithm and applications\cite{zhu2020transfer, li2017deep, mousavi2018deep, agostinelli2018reinforcement,fenjiro2018deep,arulkumaran2017deep,wang2022deep}, we focus on the DDS applications only and how DRL can benefit to its development. The two stages of DDS are discussed, including dispatching that forms service loops and routing that executes services loops.  The related literature is summarized in Table~\ref{tab:dispatching} and Table~\ref{tab:routing}.

\begin{table}[]
    \centering
    \caption{Comparison with other related surveys. Our survey propose DDS as a novel definition from a systematic level that includes multiple practical application usages. We list the main focus and the content deficiency of other survey alike compared to ours.}
    \resizebox{0.99\textwidth}{!}
    {
    \begin{tabular}{c|cc}
    \Xhline{1pt}
         Survey& Main Focus & Deficiency Compared to Ours  \\
         \Xhline{1pt}
         \cite{haydari2020deep} & Planning Problems for Transportation Systems & Logistics within DDS \\
         \cite{qin2021reinforcement}& Planning Problems in Ridesharing problems & Express and other DDS Applications\\
         \cite{yan2022reinforcement}& Planning Problems in Logistics and Supply Chain &Transportation within DDS\\
         \cite{mazyavkina2021reinforcement}& DRL for General Combinatorial Optimizations &Practical DDS Applications\\
     \Xhline{1pt}
    \end{tabular}
    }
    \label{tab:surveys}
\end{table}


Overall, this paper presents a comprehensive survey on DRL techniques for solving planning problems in DDS systems. Our contributions can be summarized as follows:
\begin{itemize}
    \item To the best of our knowledge, this is the first comprehensive survey that thoroughly defines and investigates DDS systems and up-to-date DRL techniques as solutions.
    \item We classify different stages within a complete DDS system, including the dispatching stage and the routing stage. We also investigate the common applications corresponding to the two stages, introduce the theoretical background of DRL from a broad perspective and explain several important algorithms.
    \item We investigate existing works that utilize DRL for DDS systems. We summarize these works in several dimensions and discuss the individual approaches.
    \item We illustrate several challenges and open problems in DDS using DRL. We believe the summarized research directions will benefit relevant research and help to direct future work.
\end{itemize}

The remaining survey is organized as follows. We first introduce the background of this survey, including DRL and four common DDS scenarios in Sec ~\ref{Sec_background}. The stage definition and more specific problems with corresponding solutions of both dispatching and routing are summarized in Sec~\ref{Sec_dispatching} and Sec ~\ref{Sec_routing} respectively. \zong{A few current literature of the joint modeling of the two stages are discussed in Sec 5.} The commonly used simulation environments for both stages are introduced in Sec~\ref{Sec_simulation}. Then we summarize several challenges of DRL for DDS design and open research problems in Sec~\ref{Sec_Challengs} and Sec~\ref{Sec_opportunities}. Finally, we summarize this survey in Sec~\ref{Sec_conclusion}.

    

    
    
    
    
    
    
    
    
    
    
    
    
    
    
    
    
    
    

\section{Background}
\label{Sec_background}

\begin{figure}[!htb]
\minipage{0.48\textwidth}
  \includegraphics[width=\linewidth]{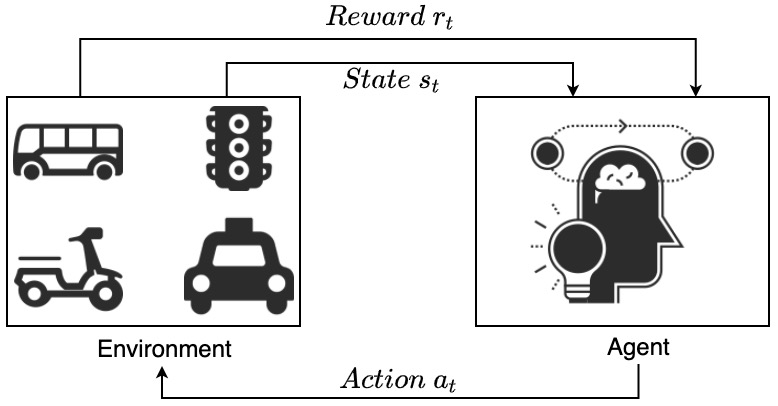}
  \caption{Reinforcement learning control loop.}\label{fig:RL_loop}
\endminipage\hfill
\minipage{0.48\textwidth}
  \includegraphics[width=\linewidth]{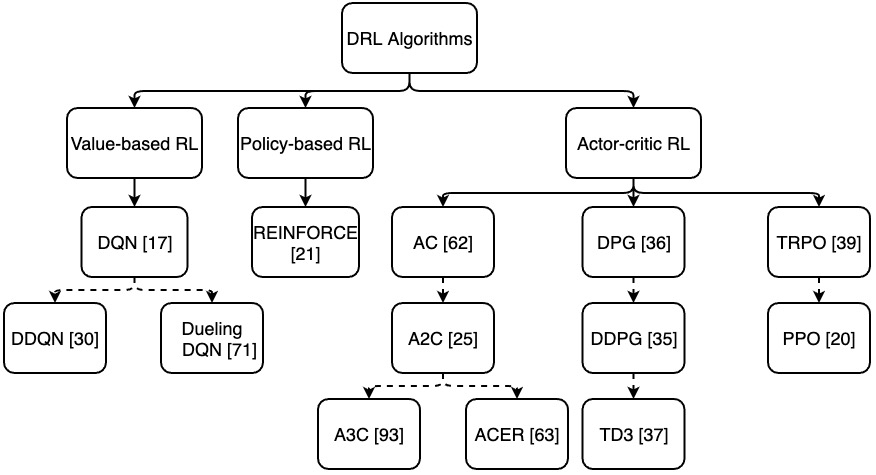}
  \caption{Classification and development of DRL algorithms.}\label{fig:drl_al}
\endminipage
\end{figure}



\subsection{Reinforcement Learning}
RL is a kind of learning that maps from environmental state to action.
The goal is to enable the agent to obtain the largest cumulative reward in the process of interacting with the environment~\cite{sutton2018reinforcement}. Usually, the Markov Decision Process (MDP) can be used to model RL problems. There are several core elements within RL under an MDP setting, including the agent, the environment, the state, the action, reward, and transition. We draw Figure 4 to represent the reinforcement learning control loop and the detailed descriptions are as follows,

\begin{itemize}
    \item \textbf{Environment.} \zong{The environment of DRL is the fundamental setting that provides basic information from exogeneous dynamics. It refers to the external system with which agents interact. The other components below can also be regarded as the sub-aspects of the environment.}
    \item \textbf{Agent.} The RL agent is supposed to provide actions and interact with the entire environment. There could be even more than one agent, which further forms the multi-agent RL setting.
    \item \textbf{State.} $S$ is the set of all environmental states. By modeling the planning task as an MDP as the prior, the state of the agent, $s_{t}\in S$, at decision step $t$ describes the latest situation. The state of the agent serves as the endogenous feature that influences the decision making. 
    \item \textbf{Action.} $A$  is the set of executable actions of the agent. The action, $a_{t}$ is the way that agents interact with the environment at decision step $t$. Any action could influence the current state. 
    \item \textbf{Reward.} $f: S\times A\rightarrow R$ is the reward function. By continuously carrying out actions to change states, the agent will finally obtain the corresponding reward, $r_{t}\sim f(s_{t},a_{t})$ that is related to the task which is obtained by the agent performing the action $a_{t}$ in the state $s_{t}$ at  decision step $t$. With $R$ as the task signal, the entire training process of RL is to obtain a high reward, which represents how successful the agent is in completing the given task.
    \item \textbf{Transition.} $p:S\times A \times S \rightarrow [0,1] $ is the state transition probability distribution function. $s_{t+1} \times p(s_{t},a_{t})$ represents the probability that the agent performs the action at in the state $s_{t}$ and transits to the next state $s_{t+1}$.
\end{itemize}

In RL, policy $\pi:S \rightarrow A$ is a mapping from state space to action space. It means that the agent selects an action with state $s_{t}$, executes the action $a_{t}$ and transits to the next state $s_{t+1}$ with probability $p(s_{t},a_{t})$, and receives the reward $r_{t}$ from environmental feedback at the same time. Assuming that the immediate reward obtained at each time step in the future must be multiplied by a discount factor $\gamma$. From the time $t$ to the end of the episode at time $T$, the cumulative reward is defined as $R_{t}=\sum_{t^{'}=t}^{T} \gamma ^{t^{'}-t}r_{t^{'}}$, where $\gamma \in [0,1]$, which is used to weigh the impact of future rewards.

The state action value function $Q^{\pi}(s, a)$ refers to the cumulative reward obtained by the agent during the process of executing action $a$ in the current state $s$ and following the strategy $\pi$ until the end of the episode, which can be expressed as $Q^{\pi}(s, a)=E[R_{t}|s_{t}=s,a_{t}=a,\pi]$. For all state-action pairs, if the expected return of one policy  $\pi^{\ast}$ is greater than or equal to the expected return of all other policies, then policy  $\pi^{\ast}$ is called the optimal strategy. There may be more than one optimal policy, but they share a state-action value function $
Q^{\ast}(s, a)=max_{\pi}E[R_{t}|s_{t}=s,a_{t}=a,\pi]$, which is called the optimal state-action value function. Such a function follows the Bellman optimality equation, $
Q^{\ast}(s, a)=E_{s^{'}\sim  S}[r+\gamma max_{a^{'}}Q(s^{'},a^{'})|s,a].
$

In traditional RL,  solving the $Q$ value function is generally through  iterating the Bellman equation
$
Q_{i+1}(s, a)=E_{s^{'}\sim  S}[r+\gamma max_{a^{'}}Q_{i}(s^{'},a^{'})|s,a],
$ 
Through continuous iteration, the state-action value function will eventually converge, thereby obtaining the optimal strategy: $\pi^{\ast}=argmax_{a \in A}Q^{\ast}(s,a)$. However, for practical problems, such a process to search for an optimal strategy is not feasible, since the computation cost of iterating the Bellman equation grows rapidly due to the large state space. 
To tackle such a problem, deep learning (DL) is introduced in RL to form deep reinforcement learning (DRL), which utilizes deep neural networks for function approximation in the traditional RL model and significantly improves the performances of many challenging applications~\cite{mnih2013playing, mnih2015human, silver2016mastering}.
In general, an RL agent can act in two ways: (1) by knowing/modeling state transition, which is called model-based RL, and (2) by interacting with the environment without modeling a transition model, which is called model-free RL. Model-free RL algorithms include two categories of algorithms: value-based methods and policy-based methods.  In value-based RL, an agent learns the value function of a state-action pair and then selects actions based on such a value function~\cite{sutton2018reinforcement}. While in policy-based RL, the action is determined by a policy network directly, which is trained by policy gradient~\cite{sutton2018reinforcement}.  We will first introduce value-based methods and policy-based methods, and then discuss the combinations of them. We drew  Figure 5 to show the classification and development of these methods. Besides, we also introduce multi-agent RL as a special category.

\textbf{Value-based RL.}
Mnih et al.~\cite{mnih2013playing,mnih2015human,silver2016mastering} first combined the convolutional neural network with the $Q$ learning~\cite{watkins1989learning} algorithm from traditional RL, and proposed the Deep Q-Network (DQN) framework. This model is first used to process visual perception, which is a pioneering and representative work in the field of value-based RL. DQN uses an experience replay mechanism~\cite{lin1992reinforcement} in the training process, and processes the transferred samples $e_{t}=(s_{t},a_{t},r_{t},s_{t+1})$ for training. At each time step $t$, the transferred samples obtained from the interaction between the agent and the environment are stored in a replay buffer $D=\{e_{1},...e_{t}\}$. During training, a small batch of transferred samples is randomly selected from $D$ each time, and the stochastic gradient descent (SGD) algorithm and TD error~\cite{sutton1988learning} is used to update the network parameters $\theta$. During training, samples are usually required to be independent of each other. Such a random sampling method greatly reduces the relevance between samples, thereby improves the stability of the algorithm. In addition to using a deep convolutional network with parameter $\theta$ to approximate the current value function, DQN also uses another network to generate the target $Q$ value. Specifically, $Q(s, a,\theta)$ represents the output of the current value network, which is used to evaluate the value function of the action pair in the current state. Meanwhile, $Q(s,a,\theta^{-})$ represents the output of the target value network, which is used to approximate the value function, namely the target $Q$ value. The parameters $\theta$ of the current value network are updated in real-time. After every $N$ iterations, the parameters of target value network $\theta^{-}$ are updated by $\theta$ and kept frozen for another $N$ iterations. The entire network is trained by minimizing the mean square error between the current $Q$ value and the target $Q$. Such a frozen target mechanism reduces the correlation between the current $Q$ value and the target $Q$ value and thus improves the stability of the training process.

The selection and evaluation of actions are based on the target value network $\theta_{i}^{-}$, which will easily overestimate the $Q$ value in the learning process.
Tackling this problem, researchers have proposed a series of methods based on DQN. Hasselt et al. ~\cite{van2016deep} proposed the Deep Double Q-Network (DDQN) algorithm based on the double Q-learning algorithm~\cite{hasselt2010double} (double Q-learning). There are two different sets of parameters in double Q-learning: $\theta$ and $\theta^{-}$. Where $\theta$ is used to select the action corresponding to the maximum $Q$ value, and $\theta^{-}$ is used to evaluate the $Q$ value of the optimal action. Such a parameter separation separates action selection and strategy evaluation so as to reduce the risk of overestimating the Q value.  Experiments show that DDQN can estimate $Q$ value more accurately than DQN.
The success of DDQN shows that reducing evaluation error on $Q$ value improves performance.
Inspired by this, Bellemare et al.~\cite{bellemare2016increasing} defined a new operator based on advantage learning (AL)~\cite{baird1999reinforcement} in the Bellman equation to increase the difference between the optimal action value and the sub-optimal action value to alleviate the evaluation error caused by the action corresponding to the largest $Q$ value. 
Experiments show that AL error terms can effectively reduce the deviation in the evaluation of the $Q$ value and promote learning quality. In addition, Wang et al.  ~\cite{wang2016dueling} improved the network architecture based on DQN, and proposed Dueling DQN that greatly accelerates learning speed.

Value-based RL methods are suitable for low-dimensional discrete action spaces. However, they cannot solve the decision-making problems in the continuous action space, such as autonomous driving, robot movement, etc. Therefore, we further introduce the policy-based RL methods that are capable of solving continuous decision-making problems.

\textbf{Policy-based RL.}
Policy-based RL~\cite{sutton1999policy} updates the policy parameters directly by computing the gradient of the cumulative reward of the policy with respect to the policy parameters, and finally converges to the optimal policy $
max_{\theta}E[R|\pi_{\theta}],
$ where $R=\sum_{t=0}^{T-1}r_{t}$ represents the sum of rewards received in an episode. The most common idea of policy gradient is to increase the probability of the  trajectories  with  higher  reward. Assume the state, action and reward trajectory of a complete episode is $\tau =\{s_{0},a_{0},r_{0},s_{1},a_{1},r_{1},...,s_{T-1},a_{T-1},r_{T-1},s_{T}\}$. Then the policy gradient is expressed as : $
g=\sum_{t=0}^{T-1}R\triangledown_{\theta}log\pi(a_{t}|s_{t};\theta).
$ Such a gradient can be used to adjust the policy parameters $
\theta\leftarrow \theta+\alpha g,
$ where $\alpha$ is the learning rate, which controls the rate of policy parameter update. The gradient term $\sum_{t=0}^{T-1}\triangledown_{\theta}log\pi(a_{t}|s_{t};\theta)$   represents the direction that can increase the probability of occurrence of trajectory $\tau$. After multiplying by the score function $R$, it can make the  probability density of the trajectories with higher rewards greater. 
While trajectories with different total rewards are collected, the above training process will guide the probability density to these trajectories with higher total rewards and maximize the corresponding appearance probability.

However, the above method lacks the ability of distinguishing trajectories with different quality, which will lead to a slow and unstable training process.
In order to solve these problems, Williams et al.~\cite{williams1992simple} proposed the REINFORCE algorithm with a baseline as a relative standard for the reward : $
g=\sum_{t=0}^{T-1}\triangledown_{\theta}log\pi(a_{t}|s_{t};\theta)(R-b),
$
where $b$ is a baseline related to the current trajectory $\tau$, which is usually set as an expected estimate of $R$ in order to reduce the variance of $R$. It can be seen that the more $R$ exceeds the reference $b$, the greater the probability that the corresponding trajectory $\tau$ will be selected. Therefore, in the DRL task of large-scale state, the policy can be parameterized by the deep neural network, and the traditional policy gradient method can be used to solve the optimal policy.

However, the policy-based RL methods are very unstable during training due to the inaccurate estimation of baseline $b$ and are inefficient due to that complete episodes are required for parameter updates. In order to solve these problems, researchers proposed  actor critic methods, which combine the value-based RL methods and policy-based RL methods.

\textbf{Actor Critic RL.}
V. R. Konda et al. ~\cite{konda2000actor} first proposed the actor-critic (AC) methods leveraging advantages from both value-based and policy-based methods.
The AC methods include two estimators:  an actor that plays the role of the policy-based method via interacting with the environment and generating actions according to the current policy, while a critic who plays the role of the value-based method by estimating the value of the current state during training. In AC methods, the critic's estimation of the value of the current state makes the RL training process more stable. In addition, there are some actor-critic RL methods introducing gradient restrictions or replay buffers so that the collected data can be reused, and therefore improve the training efficiency.

R. S. Sutton et al.~\cite{sutton2018reinforcement} proposed the Advantage Actor-Critic (A2C) method, which adds a baseline to the $Q$ value so that the feedback can be either positive or negative. V. Mnih et al. ~\cite{mnih2016asynchronous}  introduced distributed machine learning methods into A2C and got a new algorithm named Asynchronous Advantage Actor-Critic (A3C), which greatly improved the efficiency of the A2C algorithm. Wang et al. combined the AC method with experience replay and proposed actor-critic with experience replay  (ACER)~\cite{wang2016sample} method. This method enables the AC framework to train in an off-policy way to improve data utilization efficiency.  Lillicrap et al. ~\cite{lillicrap2015continuous} leveraged the idea of DQN to extend the Q learning algorithm to transform the Deterministic Policy Gradient~\cite{silver2014deterministic} (DPG) method, and proposed a Deep Deterministic Policy Gradient method (DDPG) based on the actor-critic framework, which can be used to solve decision-making problems in continuous action space. Moreover, it also introduced the replay buffer so that the collected data can be reused to improve training efficiency. Although DDPG can sometimes achieve good performance, it is still fragile in terms of hyperparameters. A common failure reason for DDPG is overestimating the real $Q$ value, thus making the learned policy worse.  To solve this problem, Fujimoto et al. ~\cite{fujimoto2018addressing} proposed Twin Delayed DDPG (TD3), which introduced three techniques based on DDPG. TD3 employs clipped double-Q learning to reduce the deviation of the $Q$ value estimation. It also utilizes delayed policy updating and target policy smoothing to reduce the impact of the $Q$ value estimation deviation on policy training. Furthermore, Schulman et al.~\cite{schulman2017proximal} 
employed the importance sampling~\cite{shelton2001importance} method and tailored the network gradient update of reinforcement learning to make the training process more robust. Schulman et al.~\cite{schulman2015trust} also proposed a method called Trust Region Policy Optimization (TRPO). The core idea of TRPO is to force the $KL$ differences of the prediction distribution of the old and new policies on the same batch of data so as to avoid excessive gradient updates and ensure the stability of the training process. However, TRPO employs the conjugate gradient algorithm to solve the constrained optimization problem, which greatly reduces the computational efficiency and increases the implementation cost. Therefore,  Schulman et al.~\cite{schulman2017proximal} proposed a Proximal Policy Optimization algorithm (PPO) to get rid of the calculations generated by constrained optimization by introducing a reduced proxy objective function. To handle more optimization scenarios with complicated constraints, Achiam et al.\cite{achiam2017constrained} proposed Constrained Policy Optimization (CPO) based on TRPO to optimize the policy within a limited policy space during each step. CPO addresses a surrogate optimization function to keep searching for safe policies.

\textbf{Multi-Agent RL.}
Many real-world problems require interaction modeling among different agents, and multi-agent RL algorithms are thus needed. A common approach is to assign each agent with a separate training mechanism. Such a distributed learning architecture reduces the implementation of learning difficulty and computational complexity. For DRL problems with large-scale state space, using the DQN algorithm instead of the Q learning algorithm to train each agent individually can construct a simple multi-agent DRL system. Tampuu et al.~\cite{tampuu2017multiagent} dynamically adjust the reward model according to different goals and proposed a DRL model in which multiple agents can cooperate and compete with each other. When faced with reasoning tasks that require multiple agents to communicate with each other, the DQN models usually fail to learn an effective strategy. To solve this problem, Foerster et al.~\cite{foerster2016learning} proposed a method called Deep Distributed Recurrent Q-Networks (DDRQN) model for multi-agent communication and cooperation with partially observable state. Except for distributed learning, other mechanisms including cooperative learning, competitive learning, and direct parameter sharing are also used in different multi-agent scenarios~\cite{bucsoniu2010multi}. Yang et al.~\cite{yang2018mean} also proposed the Mean Field Multi-agent RL (MFRL), which estimates the interaction within the agent through the average influence of the overall agent or neighboring agents.

\subsection{Application Overview}

\begin{figure*}
    \centering
    \includegraphics[width=0.85\textwidth]{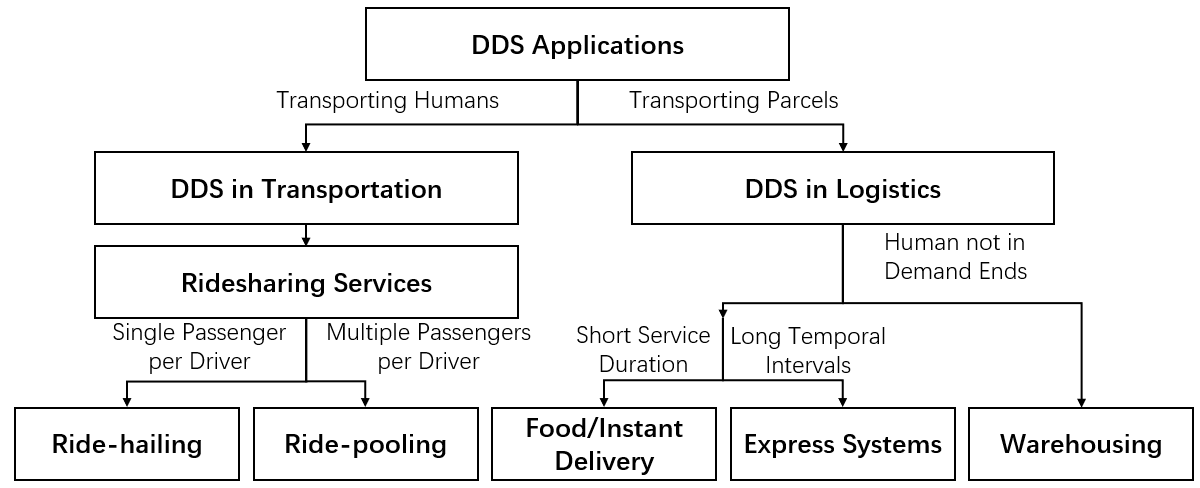}
    \vspace{-4mm}
    \caption{An overview of several urban DDS applications that have significant importance in our daily lives.}
    \label{fig:application}
    \vspace{-4mm}
\end{figure*}

As defined above, a DDS system transports either humans or parcels to provided destinations following individual demands or systematic requirements. We briefly introduce several urban DDS applications that have significant importance in our daily lives, as illustrated in Figure~\ref{fig:application}.

\subsubsection{Ridesharing}
Compared to traditional taxi-hailing services in which passengers are offered rides by chance, a ridesharing service matches passengers with drivers according to their demands from mobile apps, such as DiDi~\cite{DiDiWeb}, Uber~\cite{UberWeb}, etc. When a potential passenger submits a request from the apps to the centralized platform, the platform will first estimate the trip price and send it back. If the passenger accepts it, a matching module will attempt to match the passenger to a nearby available driver. The matching process may take time due to real-time vehicle availability and thus pre-matching cancellation may exist. After a successful match, the driver will drive to the passenger and transport him/her to the destination. A trip fare will be obtained by the driver after arrival. To reduce the average waiting time for a successful match, the platforms usually utilize a fleet management module in the backend to rebalance idling vehicles continuously by guiding vehicles to places with a higher possibility of new requests. The decisions from matching and fleet management are finally executed within the routing stage. Vehicles are navigated to serve passengers or repositioned to new areas following these strategies. In the ridesharing scenario, the service worker of a loop refers to the vehicle, while the provider and the target refer to the passengers' pickup locations and their destinations. 

A ridesharing service can be further classified into \textit{ride-hailing} where a driver is assigned only one passenger at a time, and \textit{ride-pooling} (also known as \textit{carpool}) where several passengers share a vehicle at a time. Note that in some literature the scenario of multiple passengers is also named as ridesharing. In this survey, we use ride-pooling specifically for disambiguation following ~\cite{qin2021reinforcement}.

\subsubsection{On-demand Delivery}
Many platforms around the world provide food delivery
services such as PrimeNow~\cite{PrimeNow}, UberEats~\cite{UberEats}, MeiTuan~\cite{meituan}, and Eleme~\cite{Eleme}. Except for delivering food, the newly rising instant delivery services can also deliver small parcels from one customer to another or helps to purchase other daily merchandise directly from local shops or pharmacies, such as medicines. Both food and instant delivery can be seen as a type of on-demand delivery. Compared with traditional
delivery platforms, e.g., FedEx and UPS, the orders in on-demand delivery platforms are expected to be fulfilled in a relatively short time, e.g., 30 minutes to 1 hour. A typical on-demand delivery process involves four parties: a customer as the service target, a merchant as the service provider, a courier as the worker, and the centralized platform. A customer-first places an order in a smartphone app of a platform, while a merchant starts to prepare the order and the platform assigns a courier to pick up the order. Finally, the courier delivers the order to the customer. 

\subsubsection{Express Systems}
As a long-existing DDS system, an express system is required to both pick up parcels from the consignors to the fixed depots, and deliver parcels that were loaded from the depots to the consignees. In a practical express system such as FedEx~\cite{FedEx}, Cainiao~\cite{Cainiao}, the pickup and delivery are usually considered simultaneously and could be obtained within the same service loop.  A courier loads parcels at the depot and then delivers them to their destinations one
by one via a delivery van. Meanwhile, new pick-up requests may come from local customers during the delivery
process, each of which is associated with a service
location. The courier should also go to these places to fulfill pickup requests.  Couriers are required to depart from and return to the depots by a specific time, to fit the schedule of trucks that send and pick packages to or from stations regularly.

\subsubsection{Warehousing}
Except for the DDS applications that have direct interactions with humans, the rising autonomous technologies enable unmanned management in local warehousing. The shipment requests of cargoes, usually with large size and weight, are common within a repository or among several repositories. Cargoes are moved into a targeted shelf and moved out continuously to accommodate the global shipping requirements. To reduce expenses and improve efficiency, autonomous guiding vehicles (AGVs) are commonly used in the modern warehousing scenario. In a warehousing service loop, the service provider refers to the original shelf, and the service target refers to the corresponding destination. AGVs serve as the workers in the entire process. An intelligent centralized platform is responsible to control all AGVs for efficient operations.

\begin{figure}[ht]
\vspace{-4mm}
\centering
\subfigure[On-demand Delivery.]{\includegraphics[width=0.23\textwidth,]{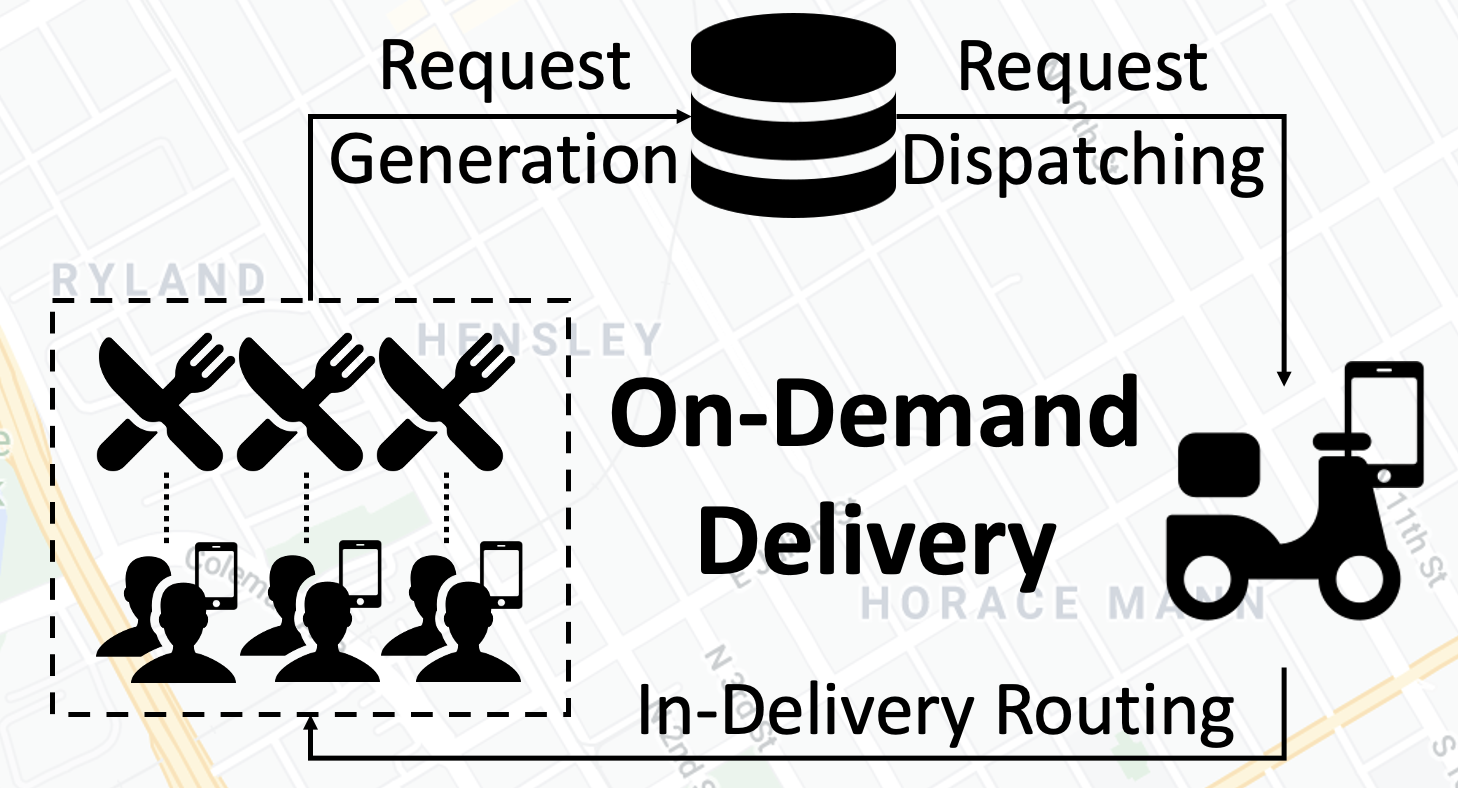}}
\subfigure[Ridesharing.]{\includegraphics[width=0.23\textwidth,]{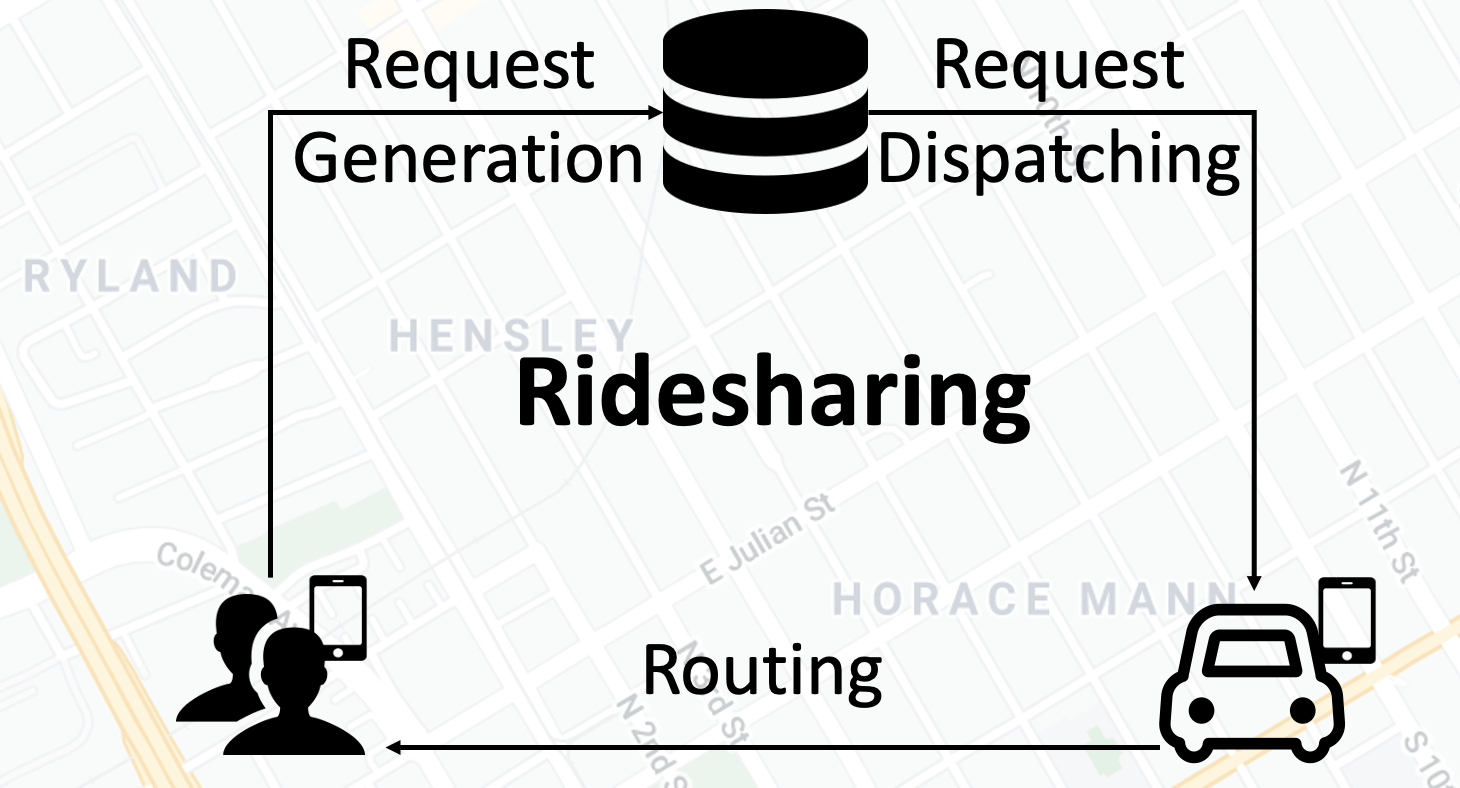}}
\subfigure[Express Systems.]{\includegraphics[width=0.23\textwidth,]{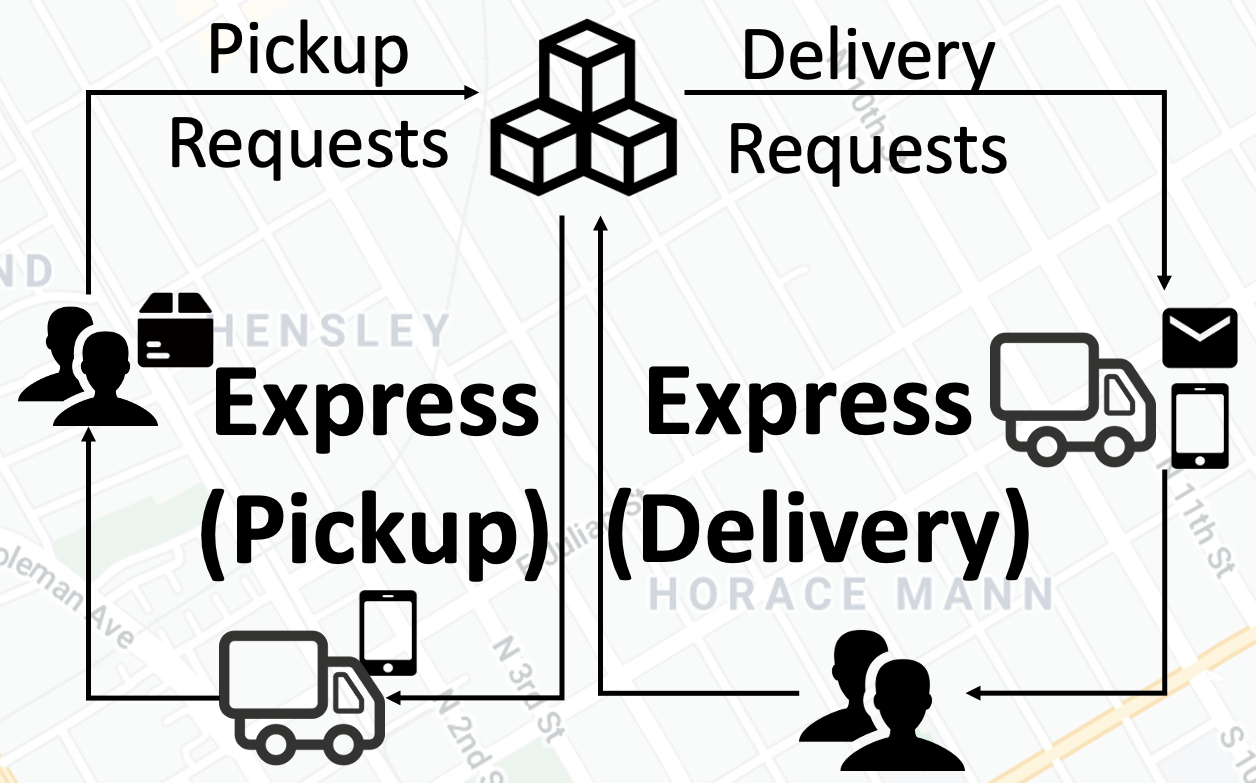}}
\subfigure[Warehousing.]{\includegraphics[width=0.23\textwidth,]{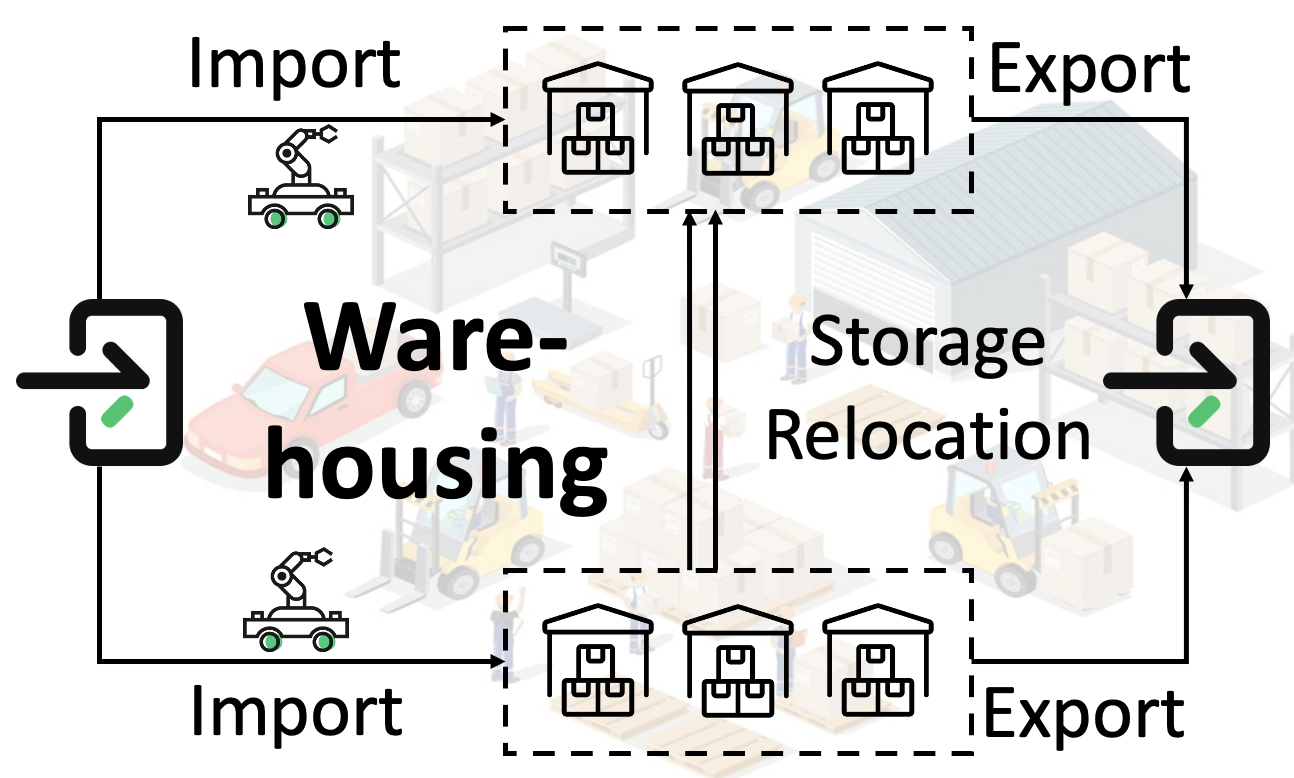}}
\vspace{-4mm}
\caption{Illustration of the four typical DDS applications.}
\label{Generalization}
\vspace{-4mm}
\end{figure}

\subsection{Relationship between Two Stages}
Generally, the research problems within practical DDS systems can be classified into two stages, i.e., dispatching and routing. The dispatching stage mainly handles the relationship between service workers and demand pairs and thus constructs service loops, while the routing stage focuses on how to execute the services within each established loop. Note that the two stages are not rigidly separated. A reasonable dispatching algorithm should consider future in-loop routing strategies as a measurement proxy. Whether a better routing solution can be generated is a direct criterion to judge different dispatching strategies. For example, a courier should not be assigned with a demand request which is far away from him since the routing distance within such a loop will be too long. On the other hand, in practical routing scenarios where a fleet of workers are on duty, it is implicated that cooperation among different workers needs consideration and thus dispatching is included. 

However, such a classification is necessary to concentrate on primary challenges in different practical scenarios. An important reference metric for such a classification we demonstrate in this survey is the \textit{Demand/Worker Ratio}. A low ratio means that the number of workers and demand pairs are balanced in each constructed loop and thus the major space of optimization is to determine how different requests should be assigned. For instance, a driver can only take one passenger in ride-hailing and no more than two passengers in ride-pooling. How to match drivers with customer requests is critical to global efficiency, while computing in-loop routing strategies is not computationally expensive. Meanwhile, in scenarios with a large ratio, it implies that a worker has to serve lots of demand requests within its loop. The routing stage, i.e., how to execute the loops thus has a high problem complexity and requires an intensive optimization process. For instance, a courier in express systems may be assigned with hundreds of parcels, and generating its optimal routing strategy becomes the primary challenge due to its NP-hard nature.

In the following sections, we will focus on the dispatching and routing stages. We will discuss the within subproblems and introduce existing solutions respectively.

\subsection{Evaluating Metrics}

We briefly introduce the major evaluating metrics of the two stages in advance.

As for the goal of dispatching, there are generally two aspects to consider, including optimizing profits for the platform and experience from the demands' side: 
\begin{itemize}
    \item \textbf{Platform Profit.} With each service loop priced, a core evaluation metric of an effective order matching system is to maximize the total revenue of all services over time. In the ride-sharing services specifically, it is also called \textit{Total Driver Income}~\cite{zhang2024nondbrem}, \textit{Accumulated Driver Income}~\cite{li2019efficient}, \textit{Gross Merchandise Volume}~\cite{xu2018large}, et al. In on-demand delivery services, \textit{Courier Profit Efficiency}~\cite{jiang2023faircod} is used to consider courier profits. While in express systems or warehousing scenarios, \textit{System Throughput} is also considered. In the fleet management problem where the dispatch distance is further considered as the operation cost, \textit{Fleet Dispatch Efficiency}~\cite{li2023integrating} is designed, positively related with profits and negatively associated with dispatching distance. Generally, the profit perspective stands for the interest of both workers and the entire platform.
    
    \item \textbf{Service Quality.} Besides maximizing the revenue from the platform's perspective, improving the service quality is also important. Since not fulfilling all demands is usual in real-world scenarios, one typical metric is to maximize the \textit{Order Response Rate (ORR)}~\cite{li2019efficient}. If overdue is permitted with penalty, then \textit{Overall Overdue Rate}~\cite{jiang2023faircod} is also calculated. 
    
    \item \textbf{Others.} Besides the fundamental metrics above, other metric might also be proposed if the optimization goal is specially designed. For instance, a metric evaluating the fairness, \textit{Variance of Courier Profit Efficiency}~\cite{shi2021learning} is designed for fair dispatching. While if the framework considers vehicle charging along with dispatching, the \textit{Availability Rate of Electric Vehicles} and \textit{Workload Rate of Chargers} are important metrics as well. \cite{zhou2023robust} further propose a \textit{Change Response Rate}~\cite{zhou2023robust} to measure the efficiency of dispatching action in response to any sudden change in demand.
\end{itemize}

As for measuring the effectiveness of a routing algorithm, researchers concern both the solution quality and the solving efficiency:
\begin{itemize}
    \item \textbf{Solution Quality.} The solution quality is the \textit{Objective\cite{chen2024efficient,jiang2024ensemble}} of the optimization itself. In the context of routing optimization, it reflects the routing \textit{Cost\cite{zong2024reinforcement}\cite{zhang2023coordinated}} directly, and can be straightly reported to be compared with other baselines.  Besides comparing the absolute length directly, reporting the \textit{Gap\cite{zong2024reinforcement}\cite{chen2024efficient}} from the optimal is another commonly adopted metric. Since the true optimal is often difficult to obtain if the routing problem is complicated, the approximate optimal is sometimes used as an alternative. 
    \item \textbf{Solving Efficiency.} To measure the solving efficiency, the \textit{Time\cite{zong2024reinforcement}\cite{chen2024efficient}} spent of obtaining the routing result also serve as an important metric. To reduce the evaluating variance, it is generally reported as the total solving time of a large number of problem instances.
\end{itemize}

\begin{table}[t]
    \centering
    \caption{\zong{Commonly adopted metrics for evaluation. We summarize them based on stage and category, as well provide specific metric name in representative literature.}}
    \resizebox{0.85\textwidth}{!}
    {
    \begin{tabular}{c|c|cc}
    \Xhline{1pt} 
        Stage& Metric Category & Specific Metric & Literature \\
    \Xhline{1pt} 
    \multirow{16}{*}{Dispatching} 
        & \multirow{8}{*}{Platform Profit}
            & Total Driver Income &\cite{zhang2024nondbrem} \\
            && Gross Merchandise Volume & \cite{xu2018large} \\
            && Accumulated Driver Income& \cite{li2019efficient}\\
            && Courier Profit Efficiency&\cite{jiang2023faircod}\\
            && Platform's Cumulative Revenue&\cite{huang2023multi}\cite{wang2024reinforcement}\\
            && Total Platform Revenue &\cite{wang2023time}\\
            && Daily Revenue&\cite{yan2023online}\\
            && System throughput &\cite{wang2016dueling}\\
            \cline{2-4}
      
        & \multirow{3}{*}{Service Quality}
            & Order Response Rate&\cite{zhang2024nondbrem}\cite{li2019efficient}\cite{wang2024reinforcement} \\
            &&Order Fulfillment Ratio &\cite{si2024vehicle}\cite{yan2023online}\\
            &&Overall Overdue Rate &\cite{jiang2023faircod}\cite{wang2023time}\\
            \cline{2-4}
        & \multirow{5}{*}{Others}
            & Variance of Courier Profit Efficiency&\cite{jiang2023faircod} \\
            && Availability Rate of Electric Vehicles&\cite{yan2023online}\\
            && Workload Rate of Chargers&\cite{yan2023online}\\
            && Change Response Rate&\cite{zhou2023robust}\\
            && Fleet Dispatch Efficiency&\cite{li2023integrating}\\
        
    \Xhline{1pt} 
    \multirow{4}{*}{Routing} 
        & \multirow{3}{*}{Solution Quality}
            & Objective&\cite{chen2024efficient}\cite{jiang2024ensemble}\cite{zhou2023towards} \\
            && Cost& \cite{zong2024reinforcement}\cite{zhang2023coordinated}\\
            && Gap& \cite{zong2024reinforcement}\cite{chen2024efficient} \\
        \cline{2-4}
        & Solving Efficiency
            & Time& \cite{zong2024reinforcement}\cite{chen2024efficient}\\
    \Xhline{1pt} 
    \end{tabular}
    }
    \label{tab:metrics}
\end{table}

\section{Stage 1: Dispatching}
\label{Sec_dispatching}
\begin{figure}
    \centering
    \includegraphics[width =0.75\textwidth]{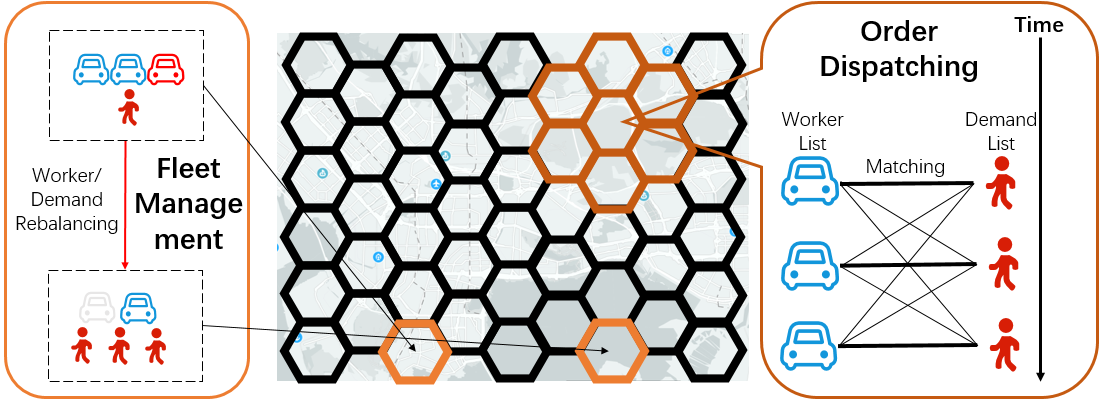}

    \caption{
    \zong{Overall dispatching architecture proposed by \cite{xu2018large}. Each hexagon here refers to a local area after discretization, while other partition schemes are also possible. In fleet management, when the workers(vehicles) are much more than the demands(passengers), the process of the management refer to repositioning some workers (in red) to a new place with more demands. While in order dispatching, the task is to match each single worker and each individual demand in detail.}
    }
    \vspace{-4mm}
    \label{fig:dispatch}
    \vspace{-1mm}
\end{figure}

Given the information of available workers and continuously updated service demand pairs, the first stage of DDS is to coordinate the relationship between demands along with available workers, and thus establish service loops efficiently. We name such a loop forming process as 'Dispatching'. Generally, the dispatching stage consists of two aspects: 1) Order matching, which aims to find the best matching strategy between workers and demands, and 2) Fleet management, which repositions idling workers to balance the local demand-supply ratio so that better order matching could be obtained in the future.  Figure\ref{fig:dispatch} shows an overview of the dispatching phase in DDS.

Formulated as an optimization problem, both tasks in the dispatching scenario are complicated due to three challenges. First, the continuously changing demand distributions and worker states bring high dynamics to the entire Markov Decision Process. It is non-trivial to accurately evaluate returns of different decision attempts. Second, a successful matching strategy should consider long-term returns\cite{zhang2017taxi}. A simple maximum result only considering current service distributions may result in a long-term loss. For example, assigning all vehicles to serve every current demand may be a local maximum in ridesharing, but may decrease the profit in the next time window since some vehicles are assigned to areas where barely any new demands appear. Third, a centralized platform should consider multiple, even large amount of workers simultaneously. Effectively modeling the cooperation and sometimes competition among them is critical to improving the system efficiency.

Concerning the given challenges, DRL has its natural advantage to solve the order matching problem compared to conventional methods and other learning-based methods. Many online reinforcement learning methods are developed to handle the non-stationariness in MDP modeling. 
Taking expected returns as learning signals, DRL is a proper framework to optimize sequential decision tasks, including dispatching tasks. Besides, modeling workers as agents is a natural way to handle the decision problem, either by homogeneously modeling all workers using the same policy, or consider the in-between interactions among multiple agents. 

In this section, we introduce both order matching and fleet management problems. Specifically for each problem, we first introduce the problem definition and common metrics, along with several conventional methods respectively. Then we thoroughly discuss detailed applications for transportation and logistics. The DRL-based literature for the dispatching stage is summarized in Tabel\ref{tab:dispatching}.

\subsection{Order Matching}

An order matching process is to assign current unserved service demands to available workers. It is also defined with other names, such as order-driver assignment in ridesharing services. The mathematical formulation originates from the online bipartite graph matching problem, where both supplies (the workers) and the demands are dynamic. It is an important module in the real-time online DDS applications with high dynamics, such as ridesharing and on-demand delivery\cite{hu2021dynamic,ozkan2020dynamic,yan2020dynamic}. Information including unserved demands, travel costs, and worker availability is updated continuously, which brings complexity to the problem.

Without purely assigning demands to workers, practical DDS systems also consider additional action choices. For instance, vehicles in a ridesharing system can be designated to idle when no proper demand can be assigned to them. As electric vehicles are widely used and deployed,  whether to recharge or continue to accept new demands forms new decision problems\cite{jindal2018optimizing}. Furthermore, controlling demand numbers assigned to a worker also expands the action space, such as considering ride-hailing and ride-pooling scenarios simultaneously\cite{jindal2018optimizing}. When each driver receives more than one customer in a loop, the action is to determine how many customers and which ones to pick up.

\subsubsection{Formulation of Online Order Matching}
Online order matching can be formulated based on a dynamic bipartite graph\cite{wang2019adaptive}, defined as $B=(L, R, E)$, where $L$ and $R$ are the sets of service workers and providers respectively and $L\cap R = \emptyset$. $E \subseteq L \times R$ is the set of edges in between. Each node $j \in R$ has its own service duration, and each possible edge $(i,j), i \in L, j \in R$ has an arrival time. The value of each edge $\omega(i,j), (i,j) \in E$ is highly related to the two values above. A matching allocation over such a dynamic graph $B$ is denoted by $M=\{(i,j)|i \in L, j \in R\}$, where each node appears at most once. The quality of $M$ can be measured as $U(B,M) = \sum_{(i,j) \in M}\omega(i,j)$. The target of online order matching is to maximize the utility score $U(B,M)$.

The optimization goal of order matching can be either optimizing profits for the platform, GMV, or the experience from the demands' size, ORR, as discussed in Section 2.4.

\begin{table*}[ht]
\centering
\caption{\zong{Literature using DRL to solve problems in the dispatching stage. The information of each literature reference consists of the publishing year, the problem solved, the application scenario, the DRL training paradigm used, the network used (if applicable), the data scheme (Ds),  the data type (Dt) used, whether the data is available (Da) and whether the code is released. OM is the abbreviation for order matching, FM for fleet management, RS for ridesharing, RP for ride-pooling, ODD for on-demand delivery, EX for express, WH for warehousing and Hex for Hexagon-grid.}
}
\vspace{-4mm}
\resizebox{0.99\textwidth}{!}
{
\begin{tabular}
{c|c|c|c|c|c|c|c|c|c}
\Xhline{1pt} 
Reference& Year & Problem &Scenario& Algorithm &Network&Ds& Dt&Da & Code \\
\Xhline{1pt} 
Li et al.\cite{li2019efficient}&2018&OM&RS& MFRL\cite{yang2018mean}&MLP&Hex/Graph&real&x&x\\
Zhou et al.\cite{zhou2019multi}&2019&OM&RS& Double-DQN\cite{van2016deep}&MLP&Hex&real, sim&\checkmark&x\\

Xu et al.\cite{xu2018large}&2018&OM&RS& TD\cite{sutton1988learning}&-&Square&real, sim&x&x\\
Wang et al.\cite{wang2018deep}&2018&OM&RS& Double-DQN\cite{van2016deep}&MLP, CNN&Hex&real&x&x\\
Tang et al.\cite{tang2019deep}&2019&OM&RS& Double-DQN\cite{van2016deep}&MLP&Hex&real&x&x\\
Jindal et al.\cite{jindal2018optimizing}&2018&OM&RP& DQN\cite{mnih2015human}&MLP&Square&real&\checkmark&x\\

He et al.\cite{he2019spatio}&2019&OM&RS& Double DQN\cite{van2016deep}&MLP, CNN&Square&real&x&x\\
Shi et al.\cite{shi2021learning}&2021&OM&RS&Value based&-&Square/Hex&real&x&x\\
Tong el al.\cite{tong2021combinatorial}&2021&OM&RS& TD\cite{sutton1988learning}&-&Square/Hex&real&x&x\\
Han et al.\cite{han2022real}&2022&OM&RS& Value based&-&Square&real&x&x\\
Eshkevari et al.\cite{sadeghi2022reinforcement}&2022&OM&RS& LM-UCB \cite{kocsis2006discounted}&-&Hex&real&x&x\\
Wang et al.\cite{wang2022fed}&2022&OM&RS&TD\cite{sutton1988learning}&-&Hex&real&x&x\\

Zhang et al.\cite{zhang2024nondbrem}
&2024&OM&RS&Q-Learning\cite{watkins1989learning}&-&Graph&real&x&x\\

Yan et al.\cite{yan2023online}& 2023&OM&RS&SARSA\cite{sutton2018reinforcement}&-&graph&real&x&x\\

Al-Abbasi et al.\cite{al2019deeppool}&2019&OM&RS& DQN\cite{mnih2015human}&CNN&Square&real&x&x\\
Qin et al.\cite{qin2021optimizing}&2021&OM&RS&AC\cite{konda2000actor}, ACER\cite{wang2016sample}&MLP&Square&real&x&x\\

Wang et al.\cite{wang2019adaptive}&2019&OM&RS&Q-Learning\cite{watkins1989learning}&-&Graph&real, sim&\checkmark&x\\
Ke et al.\cite{jintao2020learning}&2020&OM&RS&DQN\cite{mnih2015human},  A2C\cite{sutton2018reinforcement}, ACER\cite{wang2016sample}, PPO\cite{schulman2017proximal}&MLP&Square/Hex&real, sim&\checkmark&x\\
Yang et al.\cite{yang2021exploring}&2021&OM&RS&TD\cite{sutton1988learning}&MLP&Square&real&x&x\\

Si et al.\cite{si2024vehicle}&2024&OM&RS&A3C\cite{mnih2016asynchronous}&MLP,RNN&Graph&real&x&x\\

Chen et al.\cite{chen2019can}&2019&OM&ODD& PPO\cite{schulman2017proximal}&MLP&Square&real, sim&x&x\\
Wang et al.\cite{wang2023time}&2023&OM&ODD&AC\cite{konda2000actor}&MLP, GRU&Graph&real&x&x\\

Lin et al.\cite{jiang2023faircod}&2023&OM&ODD&A3C\cite{mnih2016asynchronous}&RNN&-&real&\checkmark&\checkmark\\

Li et al.\cite{li2019efficientexpress}&2019&OM&Ex&DQN\cite{mnih2015human}&MLP, CNN&Square&real&x&x\\
Li et al.\cite{li2020cooperative}&2020&OM&Ex& DQN\cite{mnih2015human}&MLP, CNN&Square&real&x&x\\

Zhang et al.\cite{zhang2023vehicle}&2023&OM&ES&Dueling DQN\cite{wang2016dueling}&MLP&Graph&sim&x&x\\

Hu et al.\cite{hu2020deep}&2020&OM&WH& DQN\cite{mnih2015human}&MLP& Graph&real&x&x\\

\hline

Lin et al.\cite{lin2018efficient}&2018&FM&RS&A2C\cite{sutton2018reinforcement}, DQN\cite{mnih2015human}&MLP&Hex&real&\checkmark&\checkmark\\

Zhang et al.\cite{zhang2020dynamic}&2020&FM&RS&Dueling DQN\cite{wang2016dueling}&MLP&Hex&real&\checkmark&\checkmark\\

Wen et al.\cite{wen2017rebalancing}&2017&FM&RS&DQN\cite{mnih2015human}&MLP&Square&real, sim&x&\checkmark\\

Oda et al.\cite{oda2018movi}&2018&FM&RS&DQN\cite{mnih2015human}&CNN&Square&real&x&x\\

Liu et al.\cite{liu2020context}&2020&FM&RS&DQN\cite{mnih2015human}&GCN&Square&real&\checkmark&\checkmark\\

Shou et al.\cite{shou2020reward}&2020&FM&RS&MFRL\cite{yang2018mean}&MLP&Square&real&\checkmark&\checkmark\\

Mao et al.\cite{mao2020dispatch}&2020&FM&RS&AC\cite{konda2000actor}&MLP&Irregular&real&\checkmark&x\\

Feng et al.\cite{feng2021scalable}&2021&FM&RS&PPO\cite{schulman2017proximal}&MLP&Square&real&x&x\\

Liu et al.\cite{liu2022deep}&2022&FM&RS&DQN\cite{mnih2015human}& MLP, CNN& Square/Hex&real&x&x\\

Huang et al.\cite{huang2023multi}&2023&FM&RS&DQN\cite{mnih2015human}&MLP&Hex&sim&x&x\\

Zhou et al.\cite{zhou2023robust}&2023&FM&RS&TD3\cite{sutton1988learning}&MLP&Square&real&\checkmark&x\\

Li et al.\cite{li2023integrating}&2023&FM&RS&DQN\cite{mnih2015human}&CNN&Square&real&\checkmark&x\\


\hline
Jin et al.\cite{jin2019coride}&2019&OM+FM&RS& DDPG\cite{lillicrap2015continuous}&MLP, RNN&Hex&real&\checkmark&\checkmark\\
Holler et al.\cite{holler2019deep}&2019&OM+FM&RS& DQN\cite{mnih2015human}, PPO\cite{schulman2017proximal}&MLP&Square-&real, sim&x&x\\

Guo et al.\cite{guo2020deep}&2020&OM+FM&RS&Double DQN\cite{van2016deep}&CNN&Square&sim&x&x\\

Liang et al.\cite{liang2021integrated}&2021&OM+FM&RS&DQN\cite{mnih2015human}, A2C\cite{sutton2018reinforcement}&MLP&Graph&real&x&x\\

Tang et al.\cite{tang2021value}&2021&OM+FM&RS&Value based&-&Hex&real&x&x\\

Sun et al.\cite{sun2022optimizing}&2022&OM+FM&RS&A2C\cite{sutton2018reinforcement}&MLP&Square&real&\checkmark&x\\

Wang et al.\cite{wang2024reinforcement}&2024&OM+FM&RS&DDPG\cite{lillicrap2015continuous}&MLP&Square&real&\checkmark&x\\

Singh et al.\cite{singh2024dispatching}&2024&OM+FM&WH&Dueling DQN\cite{wang2016dueling}&MLP&Square&real&x&x\\

\Xhline{1pt} 
\end{tabular}
}
\label{tab:dispatching}
\vspace{-4mm}
\end{table*}

\subsubsection{Conventional Methods for Order Matching}
The order matching problem and many variants were widely studied in the field of Operations Research (OR). Given the deterministic information of both workers and demands, the problem can be summarized as bipartite matching and can be solved via the traditional Khun-Munkres (KM) algorithm\cite{munkres1957algorithms}. Early methods were proposed using greedy algorithms to assign the nearest available vehicle to a ride request\cite{liao2003real}. These methods omit the global demands and supplies, and thus cannot achieve optimal performances in the long run. With new demands and worker states updating continuously, stochastic modeling becomes a major challenge. Researchers developed heuristics to deal with it efficiently\cite{ozkan2020dynamic,sungur2010model, lowalekar2018online, hu2021dynamic}. Based on historical data and the predictable pattern of demands, Sungur et al.\cite{sungur2010model} use stochastic programming to model the uncertain demands in the courier delivery scenario. Lowalekar et al.\cite{lowalekar2018online} tackle the problem with stochastic optimization with Bender's decomposition and propose a matching framework for on-demand ride-hailing. Hu and Zhou et al.\cite{hu2021dynamic} also formulate it as a dynamic problem and use heuristic policies to explore the structural space of the optimal.

\zong{
Even though these stochastic programming based methods could optimize the long-term income to some extent, the strong spatialtemporal dependency and the sequential decision nature are still challenging for the conventional methods, especially when the considered time span and objective amount are large. In comparison, DRL with explosive progress in recent years shows its much greater potential in solving stochastic and non-stationary large-scale decision problems.
}

\subsubsection{Typical Modelling of Order Matching via DRL}

\begin{figure}
    \centering
    \includegraphics[width=0.77\textwidth]{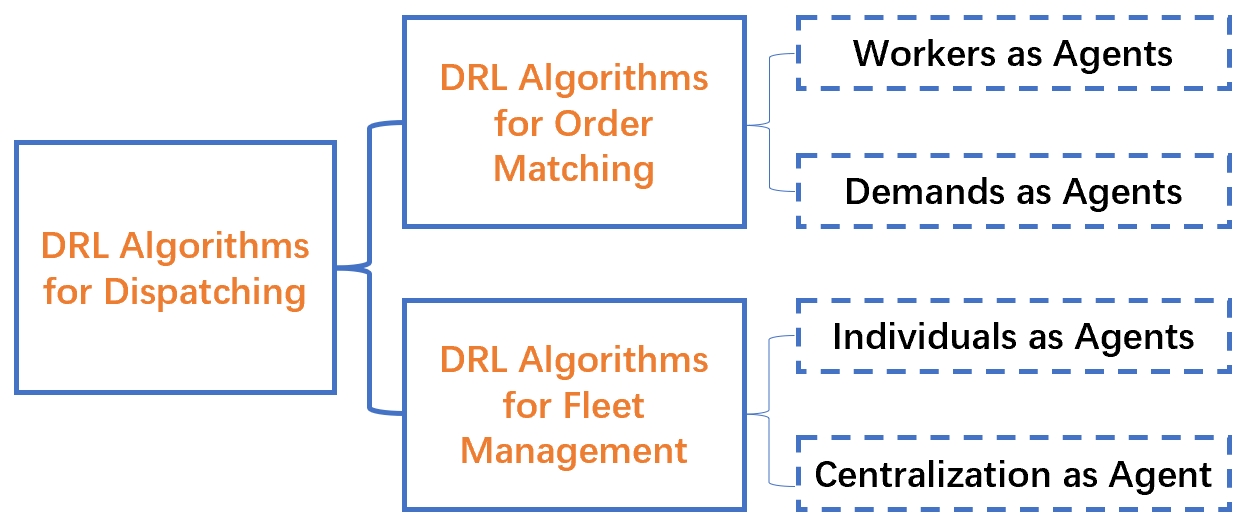}
    \caption{DRL Modeling Scheme for Dispatching problems.}
    \label{fig:dispatching_taxonomy}
\end{figure}

\zong{
In DDS, individual demand pairs are continuously generated, while a centralized platform is responsible to assign them to individual workers continuously. As a result, most DRL modeling of order matching can be classified into two perspectives as follows,
\begin{itemize}
    \item \textbf{Perspective of workers.} An intuitive way to formulate the MDP for order matching is to model all individual workers in the system as different agents. Researchers either leverage MARL techniques\cite{li2019efficient, zhou2019multi} to model the cooperation among the workers, or choose to train a single policy and implement it to all workers in a simplified way.

    \item \textbf{Perspective of demands.} Concerning the emerging demands directly, another way to formulate the MDP is to treat each request, the service providers as agents rather than the service worker. 
\end{itemize}
We show such modelling taxonomy in Fig.~\ref{fig:dispatching_taxonomy}.
 We further discuss specific literature of order matching according to the taxonomy above in both transportation and logistics systems. We summarize them in Table\ref{tab:dispatching} by listing their problem and methodology details.
}

\subsubsection{DRL for order matching in Transportation Systems}

Order matching is an essential decision and optimization problem in applications in transportation systems, such as ride-sharing services. Modern taxi and in-service vehicles can share their real-time coordinates and states to the centralized platform via mobile networks. On the other hand, each customer can generate new requests including the provided pick-up locations and the destinations as a demand pair. In DDS transportation where the demand/worker ratio is relatively low, the coordination between demands and supplies is the principal issue. 


\zong{From the perspective of workers leveraging MARL, Li et al.\cite{li2019efficient} used the Mean Field Reinforcement Learning (MFRL)\cite{yang2018mean} that models the interactions in-between each agent as the average of others.} The multi-agent framework is compared with a learning based single-agent matching algorithm and several heuristic policies considering either revenue or response.   Zhou et al.\cite{zhou2019multi} argue that no explicit cooperation or communication is needed in a large-scale scenario. They propose a decentralized execution method to dispatch orders following a joint evaluation. The framework shows advantages to heuristic methods and the single-agent modeling by Xu et al.\cite{xu2018large}. Both two algorithms considering both ADI and ORR conduct experiments based on a grid based environment and a graph-based simulator with real-world data provided by DiDi\cite{DiDiWeb}.  Even though direct multi-agent formulation in real-world scenarios are straight forward and could describe the cooperation nature in between, direct modeling without any simplification may suffer from the enormous joint action space of thousands of agents. 


As a comparison, another simplified and commonly accepted method to model the cooperation is to train a single policy and implement it to all workers online\cite{xu2018large, wang2018deep, tang2019deep, qin2020ride,tong2021combinatorial, han2022real, sadeghi2022reinforcement}. In this formulation, all workers are defined with homogeneous state, action space, and reward definitions. Even though the system is still multi-agent from the global perspective, the training stage only considers a single one. Specifically, Xu et al.\cite{xu2018large} model order matching as a sequential decision-making problem and develop a joint learning-and-planning approach. They use Temporal Difference (TD)\cite{sutton1988learning} to learn the approximate driver value function in the learning stage, and then use the KM algorithm to solve the bipartite matching problem based on learned values during planning. As an early-stage work, it is only compared with two basic matching strategies in a grid-based toy example in terms of GMV, and the design of online A/B test is introduced.  Based on such a learning-planning manner, Wang et al.\cite{wang2018deep} further augment transfer learning to increase the learning adaptability and efficiency, where the learned order matching model can be transferred to other cities. The experiments emphasized its superiority to other transfer learning methods. Tang et al.\cite{tang2019deep} further utilize the double-DQN framework to obtain a more stable learning process. Since online dynamic order matching scenario requires comprehensive consideration upon spatial-temporal features, they develop a special network structure using hierarchical coarse coding and cerebellar embedding memories for better representations. The designs of the algorithm further contribute to its performance improvement compared to \cite{wang2018deep}.  Leveraging the ST-features, He et al.\cite{he2019spatio} also develops a capsule-based network for better representations. Other than GMV and ORR, they also consider idle driving time when comparing with several model-based and heuristic methods.  Jindal et al.\cite{jindal2018optimizing} only concentrated on the ride-pooling task, and design their agent to decide whether a vehicle is to take a single or multiple passengers. Detailed matching is left to low-level algorithms. Shi et al. \cite{shi2021learning} focus on bi-objective optimization by considering fairness in addition. They propose a learning-and-planning framework following Xu et al.\cite{xu2018large}, and compare the performance with several fairness related baselines using data from three cities. Han et al.\cite{han2022real} design a fully-online dispatching system which is deployed to all cities in which Lyft operates. They factorize value function into value factors and use TD error to update values. Sadeghi et al. \cite{sadeghi2022reinforcement} also design a online RL method for ride-hailing platform and deploy it in a major international market. They design new utility function that provides more flixibility and interpretability. They also refer to the multi-arm bandit problem and introduce their LM-UCB algorithm to reduce order cancellation.

The homogeneous agent formulation avoids common challenges of multi-agent RL, including the exponential decision space of different agents. Besides, complicated communication is also avoided since all agents share the same state. Even though such a single-policy scheme is not able to handle heterogeneity and to consider individual preference, it is efficient enough in most cases.

Instead of referring different workers as agents, some researchers also treat each request from the complete request list as the agent. Yang et al.\cite{yang2021exploring}  models each demand as an agent and train a value network to estimate the values of demands instead of workers. A separate many-to-many matching process is further executed based on the learned values. They compare the framework with several non-learning based methods on a grid-based environment supported by real-world data from Hong Kong.  Since online order matching includes non-stationariness from high dynamics, some literature also attempts to find solutions by concentrating on each time window to transform it into a static problem\cite{jintao2020learning, wang2019adaptive}. Following such an agent modeling,  Ke et al.\cite{jintao2020learning} models each request as an agent, and all agents share the same policy. The action space of each agent is considered as whether to delay the current request to the next time window for further matching decisions. They compared several RL algorithms under such modeling on both a customized toy environment and a simulator based on real-world data.  Wang et al.\cite{wang2019adaptive} extends the problem scope to generalized dynamic bipartite graph matching and train a single agent which represents the entire request list and decides how long the current window lasts. They compare the framework with several online matching baselines using the open sourced GAIA dataset\cite{GAIA}. In both formulations above, eventual matching results are generated by static bipartite graph matching. 
Such a request-as-agent modeling scheme focusing on service requests or providers rather than workers benefits from handling the dynamic nature of online order matching. Since new requests keep arriving, considering requests as the agents could learn how to optimize the objectives from a global perspective. 



\subsubsection{DRL for Order Matching in Logistic Systems}
Not only important in practical applications in transportation, order matching is also essential in modern logistic systems. As pick-up requests come in real-time with many couriers picking up packages, how to manage couriers to ensure cooperation among them and to complete more pick-up tasks over a long time is important but challenging. 

With the requirement of fast responding to on-demand delivery customers, modern on-demand delivery systems need effective matching strategies to assign new demands to couriers. Chen et al. \cite{chen2019can} proposed a framework that utilizes multi-layer images of the spatial-temporal maps to capture real-time representations in the service areas. They model different couriers as multiple agents and use Proximal Policy Optimization (PPO)\cite{schulman2017proximal} to train the corresponding policy. They compare their framework with several greedy strategies and matching algorithms based on both artificial data and real-world data from Alibaba. As for the more common express systems, researchers also focus on developing an effective and efficient intelligent express system by optimizing the order matching problem.  Li et al.  proposed a soft-label clustering algorithm named BDSB to dispatch parcels to couriers in each region\cite{li2019efficient}. A novel Contextual Cooperative Reinforcement Learning (CCRL) model is further designed to guide where should each courier deliver and serve in each short period. Rather than considering both pickup and delivery tasks, Li et al. focus on pick-up tasks only and further proposed a Cooperative Multi-Agent Reinforcement Learning model to learn courier dispatching policies\cite{li2020cooperative}. Both two frameworks above are evaluated with real-world express data collected from Beijing, and compared with several non-learning and RL methods. Meanwhile, Hu et al. also develop an AGV dispatching framework using RL\cite{hu2020deep} in warehousing. Modeling each task in warehousing as an agent, the action space is consist of the scheduling rue and the AGV type to be assigned with. The framework is trained with DQN, and evaluated within a simulation environment to simulate the warehouse and the carport. 

Generally, we summarize two perspectives in modeling the order matching problems: modeling service workers as agents, either by multi-agent modeling or treating them with homogeneous policy, and modeling service providers as agents. The multi-agent modeling perspective is straight forward and could consider individual preference, but may also suffer heavy computation. Training a homogeneous policy instead could otherwise work much more efficiently, but be limited in considering heterogeneity. While modeling agents from the request side could better handle the dynamic nature since new requests keep arriving.

\subsection{Fleet Management}
When a service worker is not assigned with demands and idling temporarily, a well-considered reposition strategy upon him can increase the possibility of future service chances and thus increase the entire platform's revenue. 
Such a repositioning process forms the important fleet management problem which is also presented as vehicle positioning or taxi dispatching\cite{liu2020context, liu2023cost}. A straightforward intuition is that reasonable management can help balance the demands and supplies in different regions, thus help to improve the demand matching rate. We present the commonly accepted MDP modeling for fleet management problem and investigate the related DRL applications.

\subsubsection{Conventional Methods for Fleet Management}
Balancing the distributions of both DDS workers and demands was extensively studied, especially for the transportation systems. For instance, the balance of taxis and customers is essential in an efficient transportation system\cite{nourinejad2016developing}. Traditional methods were mostly based on data-driven approaches, which highly investigate the historical records of the supply and demand distributions. Miao et al. capture the uncertain sets of random demand probability distributions via spatial-temporal features\cite{miao2017data}. Yuan et al. and Qu et al. also construct a recommend system for vehicles to provide recommended options for repositioning\cite{qu2014cost, yuan2012t}. Various techniques, including mixed-integer programming and combinatorial optimizations, are utilized to model and solve the fleet management problem\cite{xu2018taxi, xie2018privatehunt}. 

\zong{
However, the non-deep-learning based data-driven methods cannot make full use of the historical records, since the entire modeling highly depends on handcrafted features.  Further mixed-integer programming and combinatorial optimization based methods are also facing efficiency challenges when dealing with large scaled fleet management problems. 
}

\subsubsection{DRL for Fleet Management}
Following the idea of partitioning the city area into local grids to reduce computation cost, the MDP modeling of fleet management is also constructed based on the discrete dimension space. Given the spatial-temporal states of the workers in the fleet as individual agents and the information of dynamically updated customers, an intuition is to reposition available workers to locations with a larger demand/supply ratio than their current ones. For computational efficiency, the agents within the same grid at the same period are often considered as the same agents\cite{lin2018efficient}. The goal of the platform is to maximize the long-term revenue of the entire platform of all agents or the total response rate, so as in order matching. Since measurement includes detailed matching between demands and workers, an intuitive assumption is that a worker can only be matched with the demand providers from its current neighbor grids. The action of each agent is defined based on the grid maps, which contains $x+1$ discrete action choices including moving to one of its neighbors in the $x$-way connected grids or staying as where it is. 

\zong{
The detailed method design of DRL in fleet management can also be classified into two perspectives,
\begin{itemize}
    \item \textbf{Perspective of individuals.} Since the entire fleet management process could be decomposed into the rearrangement of individual workers, an intuitive way to formulate such a problem is to learn the individual policies for the individuals.
    \item \textbf{Perspective of centralization.} Other than from the individual perspective, recent researchers also consider the system-level RL formulation, where the action space is formulated as the joint one of all vehicles.
\end{itemize}
We show such modelling taxonomy in Fig.~\ref{fig:dispatching_taxonomy}.
}

\zong{
The perspective of individuals could well address the coordination between different agents\cite{lin2018efficient, oda2018movi, wen2017rebalancing, verma2017augmenting, gao2018optimize, zhang2020dynamic, liu2020context, shou2020reward}. Lin et al.\cite{lin2018efficient} model the cooperation within the fleet as a multi-agent environment and propose a MARL-based solution for fleet management. Considering both GMV and ORR, the framework is trained by different RL algorithms and compared with several rule-based methods. Results are collected from both random dataset and a real-data based simulator. Zhang et al.\cite{zhang2020dynamic} develop a Dueling DQN\cite{wang2016dueling} based framework to learn to rewrite the current repositioning policy. All vehicles share the same policy for simplification.  Wen\cite{wen2017rebalancing} explore a new taxi driver perspective upon the fleet management problem. They focus on increasing the individual incomes of drivers and demonstrate that higher revenues for drivers can help bring more drivers into the platform, and thus improve service availability for service customers. Such an individual perspective is verified effective on a simulator based on both random data and real-world data from London. Rather than focusing on the cooperation only,  Shou et al.\cite{shou2020reward} further address the suboptimal equilibrium due to the competition among different drivers. They propose a reward design scheme as the upper level and establish multi-agent modeling of different drivers as the lower level. The framework converges in terms of ORR and overall service charge (OSC) on real-world grid-based data collected from NYC. In these works, as the action space could be extremely large for fleet management in a city, value-based RL algorithm, such as DQN\cite{mnih2015human} is commonly adopted by state-of-the-art approaches to accelerate the policy learning process. The agents could fast interact with the environment based on the learned Q-value and decide their next movements accordingly.  
}

As for the perspective of centralization, directly modeling under such a scheme faces the problem of explosive scalability, and thus requires further designs.  Mao et al.\cite{mao2020dispatch} determines the number of vehicles to reposition from one zone to another, so that the decision making is independent of total vehicle numbers. They train the framework with the AC algorithm and report the results on real-world data collected from Manhattan. The entire region is divided into irregular grids, in which vehicles are repositioned. Feng et al.\cite{feng2021scalable} propose a decomposition method that a sequence of atomic actions of vehicle repositions are optimized. The agent does not need to conduct control upon all joint-actions simultaneously, but execute them sequentially instead to reduce computational burden. The framework is trained with PPO and evaluated on real-world data collected by DiDi\cite{DiDiWeb}. Liu et al.\cite{liu2022deep} train a single-agent DQN framework with pruned global action space. They consider that local action space may fall into local optimum while full global action space is hard to explore.

In general, even though the system level modeling could consider global information by neglecting individual interactions, the scalability issue requires additional designs. As for vehicle-level modeling, diverse algorithms could be designed in terms of different optimization targets. In most cases the entire revenue of the platform is the goal and thus only cooperation requires modeling, while the competition becomes important when individual incomes are taken into account, which becomes more and more important since fairness is gradually considered in DDS services.

\subsection{Joint Scheduling of Order Matching and Fleet Management}
Besides individual studies upon order matching and fleet management, researchers also attempt to develop algorithms for both problems simultaneously\cite{jin2019coride, holler2019deep, guo2020deep, liang2021integrated, shi2021learning, sun2022optimizing}. 

Since the action spaces of the two problems are heterogeneous,  Jin et al.\cite{jin2019coride} proposed a hierarchical DRL-based structure to measure the two stages. Specifically, they design a unified action as the ranking weight vector to rank and select the specific order for matching or the destination for fleet managing. They borrow the idea from feudal networks by assigning detailed targets from manager module to worker module. The final order matching and fleet management actions are generated from the worker module. Such a joint-modeling manner shows its superiority to several RL baselines in terms of both ADI and ORR\cite{xu2018large, wang2018deep,li2019efficient}.  Holler et al.\cite{holler2019deep} separate the two phases of the joint platform by first treating the drivers as individual agents for order matching and then establishing a central fleet management agent that is responsible for all individual drivers. Rather than only consider data with same distributions, they conduct experiments on the same city environment during active hours and quiet hours. Guo et al.\cite{guo2020deep, munkres1957algorithms} use a double DQN based framework to solve the fleet management problem ahead and leave the detailed Order Matching to the traditional Khun-Munkres (KM) algorithm. Shi et al.\cite{shi2021learning} argues that the estimation of the common value function is the core of modeling both two stages. An order matching module and a fleet management module are related by a common value function, and the combined V1D3 framework outperforms previous SOTA algorithms, i.e., the winners from KDD Cup 2020 RL track competition in both sub-tasks respectively\cite{DiDi}. Sun et al.\cite{sun2022optimizing} exploit joint order matching and fleet management to optimize both efficiency and fairness. They increased both the worst and overall income.

As a conclusion, current frameworks in joint modeling both sub-tasks all utilize value-based RL algorithms to maintain a common value function to help estimate different matching and repositioning rewards. A well-designed unified latent representation of agent states is essential to augment the policy exploration ability and the robustness of training in a joint framework. A major challenge of the integrated modeling is that it is difficult to model the heterogeneous actions in two individual phases, thus the detailed action generation process is usually separated into different parts.  Further joint research of the two phases remains as the opportunity for effective DDS.

\section{Stage2: Routing}
\label{Sec_routing}

By assigning tasks to different workers and balancing the relationship between workers and demands, the service loops are constructed. The second stage of DDS scheduling is to determine how to serve each demand pair within the constructed service loops. For example, in the ride-pooling situation, a driver at a time may have several customers on the car and should decide the individual service priority. Routing is a more common problem in logistic systems where the ratio of worker/demand is much larger. For example, an express van may be assigned with more than a hundred delivery demands within its current service loop. A well-considered visiting strategy to execute the loop is critical to reducing the expenses. 

\zong{ We note that the routing stage we define here in DDS systems is different from the research series of routing planning~\cite{peng2021urban}, routing optimization~\cite{he2023routing} or vehicle navigation~\cite{sun2023hierarchical}. The objective of these works is to either find a path for the vehicle, the service worker alone, in realistic road network, or improve the entire road network load balance. Meanwhile, the \textit{Routing} problems we define here concerns the relationship between demand pairs and the service worker, where the visiting order of the demand sequence is commonly optimized. Generally, the \textit{Routing} problems can be derived from the conventional TSP and VRP problem. For convenience, we first provide a mathematical formulation of typical Capacitied VRP (CVRP), then discuss the recent DRL-based solutions on solving the routing problems. 
}

\begin{equation}
    \vcenter{\hbox{\begin{minipage}{0.40\textwidth}
    \centering
    \includegraphics[width=\textwidth]{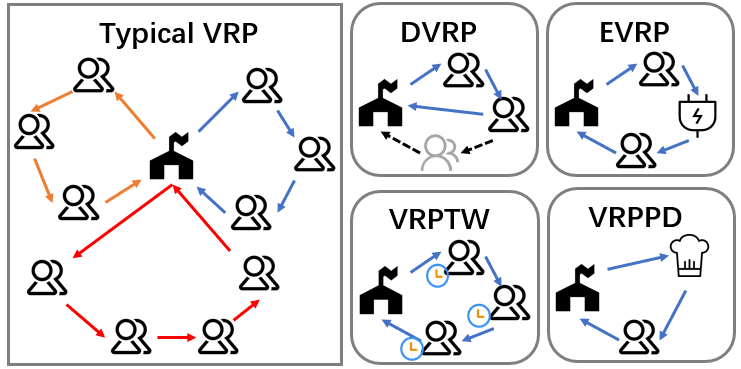}
    \captionof{figure}{A summary illustration of the typical VRP and its common variants.}
    \end{minipage}}}
\qquad
\begin{aligned}
    \label{equ:atten6}
    &\min  \ \sum\nolimits_{m=1}^{|V|}\sum\nolimits_{i=0}^{|C|}\sum\nolimits_{j=1}^{|C|+1}c_{ij}x_{ijm}, \\
    & \ s.t. \ \  \sum\nolimits_{m=1}^{|V|}\sum\nolimits_{j=1}^{|C|+1}x_{ijm}, \forall i\in C, \\
    &\quad \ \ \  \ \sum\nolimits_{i=0}^{|C|}x_{i,n+1,m}=\sum\nolimits_{j=1}^{|C|+1}x_{0,j,m}=1, \forall m \in V, \\
    &\quad \quad \ x_{ijm}\in \{0,1\}, \forall i,j \in C, \forall m in V, \\
    &\quad \ \ \  \ \sum\nolimits_{i=1}^{|C|}d_{i}\sum\nolimits_{j=1}^{|C|+1}x_{ijm} \leq c_{m}, \forall m in V,
\end{aligned}
\end{equation}

\subsection{Formulation of Typical CVRP}
The basic requirement of VRP is to design a routing strategy with a minimum cost for a fleet of vehicles, given the demands of a set of known customers.  All customers must be assigned to one vehicle to have their parcels either be picked up or delivered.  All vehicles have limited capacities, $c_{i}$, and should originate and terminate at a given depot, $v_{0}$, which also offers reloading service. 

 We represent a fleet of vehicles denoted by $V$, and a set of $n$ known customers by $C$, which formulate a directed graph $G$. The total graph includes $|C|+2$ vertices, where the depot is double represented by vertex $0$ and $n+1$. The set of arcs denoted by $A$ represents the traveling cost between customers and the depot and among customers. We associate a spatial distance $c_{ij}$ and a temporal distance cost $t_{ij}$ with each $arc(i,j)$ when $i\neq j$. $G$ includes $|V|$ subgraphs. Each connected subgraph represents a single route by vehicle $m$ and has to start from vertex $0$ and ends at vertex $n+1$ with several customers in-between, denoted by $G_{m}$. Each vehicle $m$ has a capacity $c_{m}$. Each customer $i$ has a demand $d_{i}$. The real-time shipment should not exceed $c_{m}$. 

We further denote two decision variables $x_{ijm}$ and $s_{im}$, and define $x_{ijm} = 1$, if and only if the $arc(i, j)$ is included in $G_{m}$, where $i \neq j, i\neq n+1, j \neq 0$, while $s_{im}$ represents the time stamp when vehicle $m$  serves customer $i$. By such denotations, we formulate VRP mathematically in Equation (1): 




The first three constraints make all customers visited and only visited once. and the last one indicates that a vehicle should always yield to its capacity limit.

\subsection{Realistic Routing Problems}
Besides the typical VRP setting, real-world routing problems often require additional considerations with more realistic constraints and objectives.  Many variants of VRP that tackle these practical constraints are thus closer to industrial applications and are also widely studied by researchers. We briefly introduce several important VRP variants, including dynamic VRP (DVRP), electric VRP (EVRP), VRP with time windows (VRPTW), and VRP with pickup and deliveries (VRPPD).

\subsubsection{Dynamic VRP (DVRP)}

Service demands may not be pre-obtained by the platform in real-world scenarios, thus the newly updated demands require assignment with workers dynamically~\cite{psaraftis1988dynamic}. This is the same common challenge as discussed in the dispatching stage. However, rather than simply coordinating demands with customers, the routing stage also requires a specific routing strategy with the visiting orders of the matched demands for each worker. 

Traditionally, many researchers adopt Approximate Dynamic Programming (ADP) to generate state aggregation and representation to overcome the challenge of large state space in the dynamic scenario~\cite{ulmer2019preemptive}. Rather than offline approaches, others also make online decisions to consider the relevant states at the decision point only~\cite{voccia2019same}. As for DRL based approaches, Joe et al.~\cite{joe2020deep} utilize DQN to estimate the Q-value of the individual states for vehicles and insert new demands into the existing solution sequence. The framework outperforms other non-learning baselines on 2-month logistic data. DRL is with potential to estimate the future reward with possible action attempts and is suitable to solve dynamic VRPs.

\subsubsection{Electric VRP (EVRP)}
As electric vehicles (EVs) become commonly accepted in recent years, researchers tends to study how to route the EV fleets and thus form a special Electric VRP (EVRP) problem~\cite{schneider2014electric}. They focus on the application potential of EV in both ride-hailing and express systems~\cite{shi2019operating, james2019online, zhang2022multi, zhang2021intelligent}. Since current EVs have shorter battery lives than traditional vehicles, EVRP considers the charging phase as an additional and essential action of the EV agents. Furthermore, the environment often contains information on the locations of charging stations. 

Felipe et al.~\cite{felipe2014heuristic} first propose a constructive and local search based heuristic to solve the EVRP. Goeke et al.~\cite{goeke2015routing} utilize adaptive large neighborhood search (ALNS) to solve VRP with a mixed fleet of both EVs and traditional vehicles. 
For DRL usage, Shi et al. model the EV fleet operating for ridesharing as a dynamic EVRP\cite{shi2019operating}. At each decision step, an EV agent could either pass to keep idling, to charge at the local station, or to serve the customer demands. The detailed order dispatching of customer assignment is executed by KM algorithm\cite{munkres1957algorithms}. The method is tested in a simulated environment with artificial data. 

Different from ridesharing services where the objective is to maximize the GMV, EV usage in delivery and express systems focus on how to minimize the logistic expenses. James et al. consider both charging requirements of EVs and the possibility that not all customers are visited within the given time~\cite{james2019online}. The optimization goal of such a framework is to both maximize the number of delivered logistic requests and minimize the total driving distance of all EVs. The two objectives are considered simultaneously using a weighted sum. Real-world data based simulation results show that the method could outperform conventional strategies in both static and dynamic scenarios.

\subsubsection{VRP with Time Windows (VRPTW)}
 When a service demand is provided, it may also be attached with a corresponding time window that the worker should satisfy, which means the service must arrive at the service target location within the given time window~\cite{desrosiers1984routing, solomon1987algorithms}. In practice, a customer who orders food from a restaurant may expect its food to be delivered  before it cools down. Detailed consideration of time window limits is essential in practical routing scenarios.
 
 Many heuristic algorithms were proposed to solve VRPTW, including genetic algorithm, tabu search method and ALNS~\cite{louis1999multiple, lim2004smoothed, tacs2014time}. However, it is still challenging to persuade an effective and efficient solution. 
 Zhang et al.\cite{zhang2020multi} proposed a MARL based framework by constructing the time window constraint as an additional penalty and generate the routing solutions of different vehicles one after another. All demands can thus be fulfilled even the time windows are violated. Falkner et al. \cite{falkner2020learning} proposed a joint attention mechanism to balance the coordination between vehicles and demands. Rather than only utilizing RL formulation, Zhao et al.~\cite{zhao2020hybrid} designed a hybrid structure of both DRL and local search to solve both typical VRP and VRPTW. All three approaches above conduct experiments based on simulated dataset compared with conventional methods along with benchmark RL methods, such as the AM model~\cite{ICLR19}.  When meeting all requirements is not necessary and guaranteeing time window limits is the primal goal,  James et al. \cite{james2019online} also consider the online electric vehicle routing problem, while does not force the vehicles to visit all given demands. The reward function thus consider the rate of demand response. In a word, VRPTW may vary on its detailed constraints in different scenarios, which require specific problem formulation to tackle.

\subsubsection{VRP with Pickup and Deliveries (VRPPD)}
Other than the simplified situation where the service provider and the service destination share the same location, VRPPD is a common setting in practical usage~\cite{min1989multiple}. For example, the driver for ride-sharing is supposed to first pick up the customer from the origination, and then send him to his destination. How to handle the relationship between different service provider-target pairs, i.e., pickup-delivery pairs, is of great challenge.

Previously, Li et al.~\cite{li2003metaheuristic} proposed a tabu-embedded simulated annealing approach to solve VRPPD. Ropke et al.~\cite{ropke2006unified} further utilize ALNS to tackle VRPPD. As for DRL based methods, 
Li et al.~\cite{li2021heterogeneous} \cite{li2021heterogeneous} proposes an attention-based structure by designing a special heterogeneous attention. They design several heterogeneous attention to leverage the different relations between customers within the static graph, including the pickup with paired-delivery, the pickup with other-deliveries, the pickup with other-pickups, and counter-wise if we switch the role of pickups and deliveries. Rather than a mere static scenario, Li et al.~\cite{li2021learning} consider dynamic incremental demands and propose a framework to solve the dynamic VRPPD. This also attributes to the dynamic pickup and delivery problem (DPDP) in literature. Ma et al.~\cite{DPDP} further propose a hierarchical architecture to handle when a batch of demands shall be passed to the solver and how the local batch should be solved respectively. Both two dynamic solutions outperform other conventional baselines in real-world data based experiments, with the latter one as the current state-of-the-art solution.


\subsection{Conventional Methods for Routing}
When VRP was defined in the early stage~\cite{dantzig1959truck}, researchers attempted to find exact methods to explore the exact optimal strategies.

Researchers attempted to find exact methods for solving VRP at the very beginning when it was early defined and constructed. The branch-and-bound method as a common approach for combinatorial optimizations was used as a solution~\cite{toth2002branch}. Lagrange relaxation based methods were proposed \cite{Madsen1997Vehicle, Holland1992Adaptation}, by which the problem could be solved with a minimum degree-constrained K-tree problem. Besides, Desrochers et al. firstly used the column generation to solve VRP\cite{columngeneration1992}. The following column generation based methods initialize the problem with a small subset of variables and compute a corresponding solution, and keep improving the results based on linear programming gradually. By formulating VRP as mixed integer programming (MIP), there are also several commercial solvers to generate exact solutions with small problem scales, such as Gurobi, Cplex and SCIP~\cite{bliek1u2014solving, achterberg2009scip, bixby2007gurobi}. However, due to the NP-hard nature of VRP, the performances of exact approaches are often poor and computationally expensive. The exact methods could only generate results slowly on small-sized datasets.

As a complementary to the poor performances, many heuristic-based methods were further developed to find near-optimal results instead. Compromise to the complexity of VRP and its variants, an acceptable loss on the solution quality can earn great efficiency improvement. For instance, the tabu search and local search as conventional metaheuristics were proposed to solve VRP\cite{tabuglover, groer2010library}. New solutions in the neighborhood of the current one are continuously established and evaluated. On the contrary, genetic algorithms operate in a series of solutions instead of only one solution\cite{goldberg1989genetic, Holland1992Adaptation}. Following the idea of genetics, children's solutions are generated from the best solution parents from the previous generation. Such an iteration can help to find the approximate optimal. Instead of optimizing all objectives together, ant colony optimizations utilize several ant colonies to optimize different functions: the number of vehicles, the total distance, and others\cite{macs-colony:a}. 

Even though these heuristics outperform the exact methods in finding better solutions, they are limited in real-time decision-making. For instance, genetic algorithm takes 5 hours to solve all 10000 instances of orienteering problem (OP), while the DRL based model only takes 5 seconds during inference as reported in~\cite{ICLR19}. As another drawback, the optimal approximation of the heuristic methods highly relies on manually defined rules and expert knowledge, which is far from enough compared to the enormous searching space. New technology mechanisms are needed to further improve the solution quality.



\begin{table*}[ht]
\centering
\caption{Literature using DRL to solve problems in the routing stage.  The information of each literature reference consists of the published year, the problem solved, the application scenario, the DRL training paradigm used, the network used (if applicable), the data scheme (Ds),  the data type (Dt) used, whether the data is available (Da) and whether the code is released. RS is the abbreviation for ridesharing, CVRP for capacitated VRP, DVRP for dynamic VRP, TSPR for TSP with refueling, HVRP for heterogeneous VRP, EVRP for electric VRP, VRPTW for VRP with time windows, PDP for pickup and delivery problem, WH for warehousing.
}
\resizebox{0.99\textwidth}{!}
{
\begin{tabular}{c|c|c|c|c|c|c|c|c|c}
\Xhline{1pt} 
Reference& Year & Problem &Scenario& Algorithm &Network&Ds& Dt&Da & Code \\
\Xhline{1pt} 

Nazari et al. \cite{NIPS18-VRP}&2018&CVRP&Math&PG~\cite{williams1992simple}, A3C\cite{mnih2016asynchronous}&RNN&Graph&sim&\checkmark&\checkmark\\
Kool et al. \cite{ICLR19} & 2019 &CVRP&Math&PG\cite{williams1992simple}&ATT&Graph&sim&\checkmark&\checkmark\\

Chen et al. \cite{chen2019learning}&2019&CVRP&Math&A2C\cite{sutton2018reinforcement}&MLP&Graph&sim&\checkmark&\checkmark\\
Lu et al. \cite{ICLR20}&2019&CVRP&Math&PG~\cite{williams1992simple}&MLP, ATT &Graph&sim&\checkmark&\checkmark\\
Duan et al. \cite{duan2020efficiently}&2020&CVRP&Logistics&PG\cite{williams1992simple}&GCN, ATT&Graph&real&x&x\\
Delarue et al. \cite{delarue2020reinforcement}&2020&CVRP&Math&Model-based&MLP&Graph&sim&x&x\\
Xin et al. \cite{xin2020multi}&2020&CVRP&Math&PG\cite{williams1992simple}&ATT&Graph&sim&x&x\\
Kwon et al.\cite{kwon2020pomo} &2020&CVRP&Math&PG\cite{williams1992simple}&ATT&Graph&sim&\checkmark&\checkmark\\
Ma et al. \cite{ma2021learning} &2021&CVRP&Math&PPO\cite{schulman2017proximal}&ATT&Graph&sim&\checkmark&\checkmark\\
Hottung et al. \cite{hottung2021efficient}&2022&CVRP&Math&PG\cite{williams1992simple}&ATT&Graph&sim&\checkmark&\checkmark\\

Zhou et al.\cite{zhou2023towards}&2023&CVRP&Math&PG\cite{williams1992simple}&ATT&Graph&sim&\checkmark&\checkmark\\

Chen et al.\cite{chen2024efficient}&2024&CVRP&Math&PG\cite{williams1992simple}&ATT&Graph&sim&\checkmark&\checkmark\\

Jiang et al.\cite{jiang2024ensemble}&2024&CVRP&Math&PG\cite{williams1992simple}&ATT&Graph&sim&\checkmark&\checkmark\\

Ma et al.\cite{ma2024learning}&2024&CVRP&Math&PG\cite{williams1992simple}&MLP,GRU&Graph&sim&\checkmark&\checkmark\\

\hline
Joe et al. \cite{joe2020deep}&2020&DVRP&Logistics&DQN\cite{mnih2015human}&MLP&Graph&real&x&x\\

Pan et al.\cite{pan2023deep}&2023&DVRP&Logistics&A3C~\cite{mnih2016asynchronous}&RNN&sim&real&x&x\\
Ottoni et al. \cite{ottoni2021reinforcement}&2021&TSPR&Math&Q-Learning~\cite{watkins1989learning}, SARSA~\cite{sutton2018reinforcement}&-&Graph&sim&x&x\\
Qin et al.\cite{qin2021novel}&2021&HVRP&Math&Double-DQN\cite{van2016deep}&MLP, CNN&Graph&sim&x&x\\
Bogyrbayeva et al.\cite{bogyrbayeva2021reinforcement}&2021&EVRP&RS&PG\cite{williams1992simple}&RNN&Graph&sim&x&x\\

Shi et al.\cite{shi2019operating}&2020&DEVRP&RS&TD\cite{sutton1988learning}&MLP&Graph&sim&x&x\\
James et al.\cite{james2019online}&2019&EVRPTW&Logistics&PG\cite{williams1992simple}&RNN&Graph&real&x&\checkmark\\

Lin et al. \cite{lin2020deep}&2020&EVRPTW&Logistics& PG\cite{williams1992simple}&ATT, RNN&Graph&sim&x&x\\

Chen et al.\cite{chen2022deep}&2022&EVRPTW&Logistics&PG\cite{williams1992simple}&ATT,GRU&Graph&real&x&x\\

Zhang et al. \cite{zhang2020multi}&2020&VRPTW&Logistics&PG\cite{williams1992simple}&ATT&Graph&sim&x&x\\

Falkner et al.\cite{falkner2020learning}&2020&VRPTW&Math&PG\cite{williams1992simple}&MLP, ATT&Graph&sim&\checkmark&x\\

Zhao et al. \cite{zhao2020hybrid}&2020&VRPTW&Math&AC~\cite{konda2000actor}&ATT&Graph&sim&\checkmark&x\\

Zhang et al.\cite{zhang2023coordinated}&2023&VRPTW&Logistics&MARL&ATT&Graph&sim&x&x\\

Zong et al.\cite{zong2024reinforcement}&2024&VRPTW&Logistics&PG~\cite{williams1992simple}&ATT&Graph&sim,real&x&x\\

Li et al. \cite{li2021heterogeneous}&2021&PDP&Math&PG\cite{williams1992simple}&ATT&Graph&sim&x&x\\

Chen et al.\cite{chen2023reinforcement}&2023&PDP&Logistics&PG\cite{williams1992simple}&GNN&Graph&sim,real&x&x\\

Li et al.\cite{li2021learning}&2021&DPDP&Logistics&Double-DQN\cite{van2016deep}&MLP, ATT&Graph&real&x&x\\

Ma et al.\cite{DPDP}&2021&DPDP&Logistics&DQN~\cite{mnih2015human}, PG~\cite{williams1992simple}&MLP, ATT, GNN&Graph&real&x&x\\

Lee et al.\cite{lee2021mobile}&2021&PDP&WH&Q-Learning\cite{watkins1989learning}&-&Square&sim&x&x\\

\Xhline{1pt} 
\end{tabular}
}
\label{tab:routing}
\end{table*}

\subsection{DRL for Routing}


\zong{
In recent years, many researchers attempt to utilize the deep reinforcement learning method to solve VRP and other combinatorial optimizations due to the ability to improve solution quality via a self-driven mechanism and the potential of an efficient solution generation process.} With solution quality guaranteed, DRL-based methods benefit from the separation of offline training and online inferring. Even though it may take hours or even days to train a fully converged policy in the offline stage, the inference to new problem instances in industrial online applications may only take a second compared to the meta-heuristics that will take minutes or hours~\cite{ICLR19}.  
\zong{
Generally, current works using DRL for solving VRP and its variants can be classified into sequence generation based methods and rewriting based methods. We introduce the detailed modeling and latest literature of each.
}

\subsubsection{Sequence Generation Methods}

\begin{figure}[t]
\minipage{0.48\textwidth}
  \includegraphics[width=\linewidth]{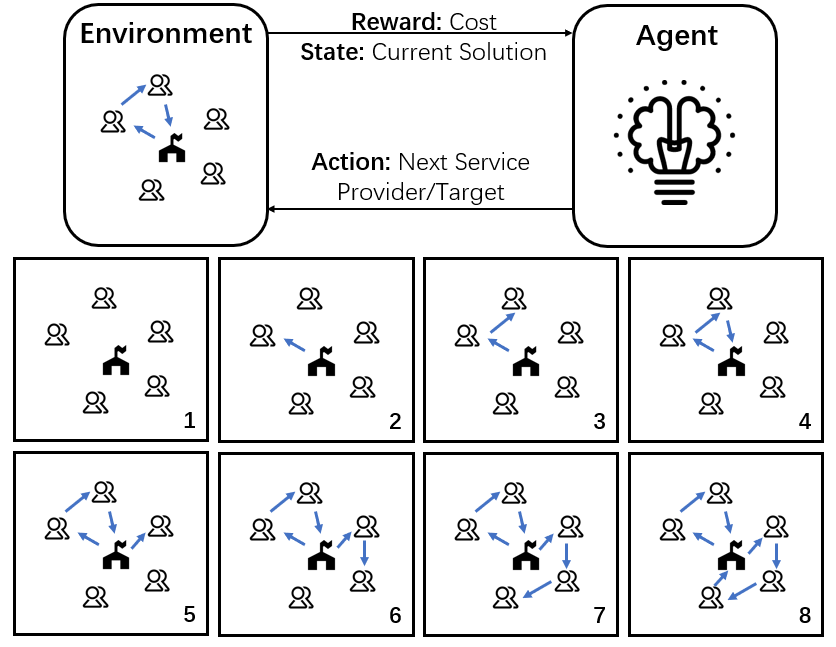}
  \caption{The illustration of sequence generation methods for generating VRP solutions via RL. In each decision step, the agent selects the next provider/target location to visit.}\label{fig:sequence}
\endminipage\hfill
\minipage{0.48\textwidth}
  \includegraphics[width=\linewidth]{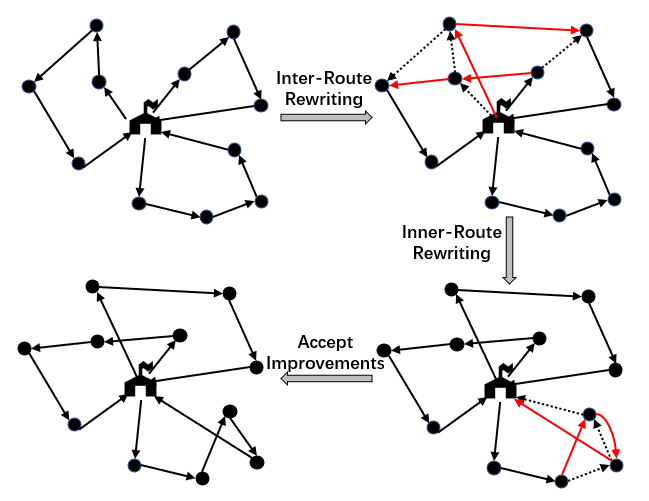}
  \caption{The illustration of rewriting based methods for generating VRP solutions via RL. An initial solution is pre-established, and the agent continuous optimizing the current solution.}\label{fig:rewriting}
\endminipage
\end{figure}

\textbf{Typical Modelling of Sequence Generation Methods.}
A common approach to generate VRP solutions is to generate a partial sequence gradually and finally obtain a complete solution. In such an MDP modeling, the updating between different states is to include a new unvisited node into the current solution, which naturally forms the action of the sequence generation agent. Taking Kool et al.~\cite{ICLR19} as an example, a policy network is constructed via an encoder and a decoder. The encoder utilizes the attention mechanism to capture the relationship between different nodes, while the decoder further calculates the probability of selecting each nodes as the policy output.  In each decision step, agent samples such a probability distribution and further selects one node to visit for exploration. The reward is designed as the negative of solution length, so that the travel cost could be minimized. It is notable that in more realistic VRP variants, practical constraints may limit the selection space since unfeasible solutions are possible to be generated. The agent thus should consider these constraints, a mask scheme is also designed to filter out the unfeasible choices. The sequence generation method is illustrated in Figure~\ref{fig:sequence}.

\textbf{Existing Literature.}
A special Pointer Network was first proposed under such a mechanism. Following a classic encoder-decoder structure, the PN  structure is independent of the encoder length, and the output sequence is a subset of the input with a generated order~\cite{vinyals2015pointer}. Even though the original PN was trained using supervised learning, it started following research exploration via DRL for more advanced VRP solutions. The basic structure of it is commonly utilized in the following research for routing problems. Bello et al.~\cite{ICLR17} first developed a special neural combinatorial optimization framework (NCO) via DRL, which showed its effectiveness on both performance and generating efficiency on TSP and the knapsack problem. Even though the typical VRP was not studied, NCO serves as an important benchmark that utilizes DRL to explore more effective combinatorial optimization solution structures. Nazari et al.~\cite{NIPS18-VRP} further followed the NCO structure and first applied it to VRP.  Kool et al.~\cite{ICLR19} combined the structure with attention mechanism as augmentation and obtained performance improvement. They investigated several routing formulations, including the TSP, typical CVRP, Split Delivery VRP (SDVRP), Orienteering Problem (OP), Prize Collecting TSP (PCTSP), and Stochastic PCTSP (SPCTSP). The attention-based structure was further developed by the following researchers. Rather than relying on a single decoder for sequence generation, Xin et al.\cite{xin2020multi} proposed a multi-decoder mechanism to generate several partial solutions simultaneously and combine them using tree search to expand the searching space. Duan et al.\cite{duan2020efficiently} on the other hand, focused on more effective feature representation ability of the network itself. They augmented the structure with GCN, and develop a joint learning approach using both DRL and supervised learning.  They evaluated the framework upon both generated random dataset as previous methods above used and real-world logistic dataset collected by Cainiao~\cite{Cainiao}. Kwon et al.\cite{kwon2020pomo} suggested to explore for solutions starting from several different nodes instead of one in for each instance, and can such explore multiple local optima simultaneously. Hottung et al.\cite{hottung2021efficient} proposed a new solution searching mechanism called active search, which updates the parameters of the neural network with a guided signal during the inference stage to explore for more optimal solutions.Zong et al.\cite{zong2022mapdp} further tackles pickup-delivery problem with multiple vehicle cooperation. \zong{Recent research further explores how to generalize the model in the presenece of a distribution shift. \cite{jiang2024ensemble} propose an ensemble-based DRL framework by learning a group of sub-policies. Furthermore, \cite{zhou2023towards} considers the generalization issue across both distribution and size by training a meta-learning framework. \cite{chen2024efficient} also adopts the meta-DRL technique to further solve the multi-objective routing problems.}

Since training and inference are separated, the sequence generation methods benefit from high inference speed. For instance, generating routing solutions for 100 customers takes only 8 seconds, while the state-of-the-art heuristic-based solver, LKH3\cite{helsgaun2017extension}, takes more than 13 hours as a comparison. Sequence generation-based methods have great potential in handling new demand changes in online platform implementation. When new DDS demands arrive or past ones are either modified or canceled, such a fast framework could make real-time responses to these changes. 

\subsubsection{Rewriting based Methods}

\textbf{Typical Modelling of Rewriting based Methods.}
Besides generating partial solutions until completeness, researchers also searched for other MDP formulations for solving VRPs. An intuition originates from the continuous modification of current solutions in operations research (OR), which is the core idea of many practical heuristics for VRP. Following this idea, researchers attempted to parameterize such a modification process to continuously improve solution quality. Such a modification process is also called Rewriting-based methods~\cite{chen2019learning}. Taking ~\cite{ma2021learning} as an example, a given operator, such as swapping the positions of two nodes, is executed continuously to update the current solution. A policy network is thus trained to calculate the probability of executing such an operator at a given position. When the position pair is sampled, the operator thus rewrites the solution into a new one. The reward is designed as the negative of solution length, so that the travel cost could be minimized. The rewriting-based method is illustrated in Figure~\ref{fig:rewriting}.

\textbf{Existing Literature.}
A framework is proposed by Chen et al.\cite{chen2019learning} where a complete solution is constructed at the first stage and promoted gradually guided by the RL agent. A local rewriting rule is designed that keeps updating the current solutions. Lu et al.~\cite{ICLR20} further propose a Learn-to-Iterate (L2I) framework which not only improves the current solution but also creates perturbation for more exploitation choices. They construct an operation pool from which the agent selects operations continuously. Results on random dataset following previous works demonstrate its potential. It is the first RL based method that outperforms the famous heuristic baseline LKH3~\cite{helsgaun2017extension} and is the current state-of-the-art method. Further realistic routing solutions also verifies the effectiveness of such a rewriting manner by combining RL and OR, such as solving VRPTW by combining local search techniques~\cite{zhao2020hybrid}. Ma et al.\cite{ma2021learning} learns to optimize the usage of a fixed operator by selecting the optimal operation location on each solution, and obtained the state-of-the-art performances. They train the agent to efficiently execute the operator via PPO.  Liu et al.\cite{liu2023neurocrossover} further proposed to conduct a neural crossover process to search for more effective solutions. 

\zong{The rewriting based methods are relatively slower, since a rewriting process requires operation duration. However, they show great potential in reaching the true optimal and outperforming the state-of-the-art conventional methods in solution quality.}


\section{Joint Modeling of the Two Stages}
Merely modeling and generating solutions for one stage in DDS is often not enough to formulate complicated real-world scenarios. As a result, researchers adopt joint-modeling for more practical DDS frameworks. From the dispatching perspective, joint-modeling is necessary when the \textit{Demand/Worker Ratio} increases, meaning that detailed visiting orders of each worker require specific optimization. Chen et al.~\cite{chen2019can} argue that in a courier dispatching problem, the detailed vehicle routing module of each couriers could be formulated as the DVRP with time windows due to the dynamic demands. Meanwhile, from the routing perspective, DVRP can also be solved from a myopic manner by formulating a dispatching process, as shown by Shi et al.~\cite{shi2019operating}. The action space for solving the original dynamic EVRP is designed as whether to appoint a vehicle to pass, charge or assign with a customer. The further routing process in the application is neglected by setting the 'first-come-first-serve' rule. In a word, dynamic DDS with relatively large \textit{Demand/Worker Ratio} requires consideration from  both dispatching and routing stages.

As discussed above, a common approach to such joint-modeling is to first match different workers and service pairs every time new service pair arrives, where dispatching (mainly order matching) techniques could be used. Then the problem could be decomposed into multiple routing problems, each consist of only one service worker with several pairs. Shi et al.\cite{shi2019operating} focus on the ridesharing service by electric vehicles, and adopts RL in the dispatching stage, while leaving the routing stage to the 'first-come-first-serve' rule as discussed above. Li et al.\cite{li2021learning} focus on the industrial logistics with a similar modeling scheme, while leaving the routing stage to a greedy strategy. The same RL-for-dispatching and rule-for-routing structure is also adopted by Shah et al. ~\cite{shah2020neural} in ride-pooling.

As an alternative, Ma et al.~\cite{DPDP} propose an bi-level RL based framework for both dispatching and routing for logistics. Rather than model each courier as an agent, they consider the request list as the agent for the upper level structure as discussed in Sec 3.1.3. Such an upper level agent is responsible for dividing and releasing batches consist of accumulated demands. Detailed solutions for static multi-vehicle routing are generated by a lower level agent. However, there is not many solutions adopting such a double RL framework for both stages, \zong{and it still remains as an open research problem, which will be further discussed in Sec~\ref{Sec_opportunities}}.
\section{Open Simulators and Datasets for DDS}


\label{Sec_simulation}
Since the existing RL methods for DDS problems are model-free, a large amount of training data generated by interaction with the environment is required. However, direct interactions with the real environment indicate high costs and high risks. Therefore, simulating upon DDS scenarios is very necessary. A reliable simulator is of great practical significance. There are already some existing DDS simulators. We will introduce existing open simulators and datasets.

\textbf{Simulator and dataset  for Dispatching.}

Simulators for dispatching learn orders' generation and state transition from the real data~\cite{tang2019deep}. There are many public dispatching-related datasets. The most commonly used data set is provided by the New York City TLC (Taxi and Limousine Commission) ~\cite{TLC}, which contains travel records for various services, Yellow taxis, Green taxis, and FHV (For-Hire Vehicle) from 2009 to 2020\footnote{\zong{https://www1.nyc.gov/site/tlc/about/tlc-trip-record-data.page}}. \cite{Kaggle} provides a subset of NYC FHV data, which contains GPS coordinates for pickup locations\footnote{\zong{https://www.kaggle.com/datasets/fivethirtyeight/uber-pickups-in-new-york-city}}. Travel time data between OD pairs can be obtained through Uber Movement ~\cite{Uber}. In addition, Didi Chuxing has released the travel records (regular and aggregated) and vehicle trajectory datasets of Chengdu, China through~\cite{GAIA}, from which they also developed a simulator to model the dispatching state\footnote{\zong{https://outreach.didichuxing.com/}}. The simulator usually consists of two parts: order generation model, driver movement, and transition model. The order generation model learns the order generation and distribution, while the driver movement and transition model learn the state transition from the dataset. The simulator for dispatching often verifies the real effectiveness of the simulator by comparing the gross merchandise volume (GMV) generated by its simulation with the GMV of real data~\cite{lin2018efficient}. To make simulators more realistic, the climate and traffic conditions are also modeled~\cite{li2018dynamic}. Recently, DiDi ~\cite{DiDi} has developed its dispatching simulation platform based on the research of existing dispatching simulators, which serves as an open-source ride-hailing environment for training dispatching agents\footnote{\zong{https://www.biendata.xyz/competition/kdd\_didi/}}.

\textbf{Simulator and dataset for Routing.}

Simulator for routing learns the orders' and vehicles' generation and state transition from historical data~\cite{li2021learning}. When the simulation starts, it first initializes orders' and vehicles' states. The RL agent observes the state and then dispatches a determined vehicle to serve an order. After the RL agent implements the learned policy, the generation model and vehicle state transition model are utilized to update the selected vehicle’s information. This process repeats until the end of the simulation.  Recently, Huawei~\cite{huawei} has developed its simulation platform based on the research of existing routing simulators,  which serves as an open-source vehicle routing environment for training dispatching agents to tackle the scenario in logistics based on dynamic VRP with pickup and delivery\footnote{\zong{https://competition.huaweicloud.com/information/1000041411/introduction}}. As for open datasets for routing, ~\cite{CVRPLib} summarized several open CVRP instance datasets with different scales with maximal to $30000$\footnote{\zong{http://vrp.atd-lab.inf.puc-rio.br/index.php/en/}}. VRPTW as a special variant routing problem has its public dataset included in  Solomon dataset~\cite{Solomon-benchmark}\footnote{\zong{https://www.sintef.no/Projectweb/TOP/VRPTW/Solomon-benchmark/}} and Homberg dataset~\cite{Homeberg-benchmark}\footnote{\zong{https://www.sintef.no/projectweb/top/vrptw/homberger-benchmark/}}. Meanwhile, the Li\&Lim benchmark~\cite{LiLim-benchmark} tackles VRPPD with time windows specifically\footnote{\zong{https://www.sintef.no/projectweb/top/pdptw/}}.

\section{Challenges}
\label{Sec_Challengs}

Owing to the advantages of DRL, many practical frameworks to tackle the two stages for DDS are developed and could generate high-quality solutions efficiently on large scales. However, challenges still exist in building more practical DDS applications. We briefly summarize the major challenges in developing DDS solutions.

\subsection{Coupled Spatial-Temporal Representations}
Capturing the dynamic changes of distributions within different service loops is essential to an effective DDS application. A good ST-representation can well reflect the in-between spatial relationships and the potential temporal consumption to accomplish the services. For instance, He et al.~\cite{he2019spatio} develop a capsule-based network to capture the representations of both new demands from passengers and available drivers in the order matching task. A well-learned representation can enhance the performance of the overall framework.

Even though ST-representation is not a new research topic and has been addressed as an important task in urban computation literature~\cite{zong2019deepdpm,feng2018deepmove,lin2019deepstn+}, the coupled ST forms a new challenge. In most DDS applications, a service target is a bond with its provider, thus the ST representation should reflect such a paired relationship. Some literature proposes special designs for the coupled challenge. For instance, Li et al.~\cite{li2021heterogeneous} propose a special attention-based structure to leverage different relationships among all customer nodes in VRP with pick and delivery. Six different attention mechanisms in total are computed as a thorough measurement upon all nodes. However, the above solution follows an ergodic way to consider all possible solutions in all extents of measurement based on the given network structure. Developing more eligible and flexible representation methods, learning mechanisms and overall algorithms are still challenges for DDS development.



\subsection{Safety of DDS Implementation}
Uncontrollable behaviors during agent model inference and unsatisfactory explainability limits the safety during DDS implementation. In real-world applications, implementation of an already-trained model should be further adapted to new unseen scenarios, while it may result in violation of multiple practical constraints.
Considering multiple constraints in DRL designs is an important research problem that has been widely studied~\cite{geibel2006reinforcement, miryoosefi2019reinforcement}. Meanwhile, the practical constraints in DDS design are especially significant.  For instance, a practical routing problem is much more complicated than the mathematical VRP due to numerous constraints, including time windows, charging requirements, structural limitations between pickups and deliveries. Effective DRL training requires corresponding considerations on these additional constraints.

A commonly used solution is to develop ad-hoc designs to these constraints. 
However, adopting such a solution for all constraints may result in complicated structure design when the constraint amount increases. It may be much more difficult to train the entire model. Another promising direction is to utilize safe RL techniques to handle the safety challenge. For instance, to deal with VPR with time windows, Zhang et al.~\cite{zhang2020multi} measure possible violation on time windows as a penalty to the total reward. The agent can learn to minimize the exceeds upon the constraint. Besides such a reward shaping method. \cite{tang2022learning} utilizes Lagarangian Relaxation to solve constrained routing problems. However, current results cannot achieve a significant enough performance, and the safety of DDS systems remains a great challenge.

\subsection{Large-Scale Deployment}
Implementing algorithms on large scales is a necessary step from pure research to industrial usage. However, training models directly in large-scale scenarios requires enormous computation resources and time. To tackle such a challenge, a commonly used method is to abandon the natural formulation of multi-agent on different service workers. Either directly providing centralized control or modeling them as homogeneous agents with shared parameters can help to simplify the training process~\cite{xu2018large,wang2018deep,tang2019deep}. Another approach of reducing computation burden is to utilize the idea of divide and conquer by reducing a city-wide planning task into multiple regional ones~\cite{li2019efficient, zong2022rbg}. Such an idea is widely used in real-world on-demand delivery systems, where the entire delivery scope is divided into regions and couriers are assigned to accomplish "last kilometer deliveries"~\cite{taniguchi2014city}.

However, current solutions are still far from enough to solve the large-scale problem, especially in the routing stage. With an NP-hard nature, the complexity of generating an optimal solution grows exponentially along with the problem scales. As a result, most existing literature on solving routing problems limit their experiment scales to no more than one hundred demand nodes within a graph-based data scheme~\cite{ICLR19, ICLR20}. Developing new training frameworks via either more agile formulation or more advanced lightweight training algorithms can help to fit in large-scale environment and promote deployment ability.

\subsection{Dynamics and Real-time Scheduling}
Real-world DDS scenarios include high dynamics from the environment. New demands arrive continuously and the existing ones may also change. For instance, a passenger who calls for a ride-hailing service may change his destination or even cancel the current request directly. Such dynamics are critical in real-time scheduling in both stages.

Much existing literature using DRL for dispatching measures the dynamic features explicitly using specially designed network structures. For instance, Tang et al.~\cite{li2021learning} represent the spatial-temporal features using hierarchical coarse coding, and He et al.~\cite{he2019spatio} develop a special capsule-based network accordingly. As for the routing stage, DVRP is specially constructed to leverage the dynamics in real-time scheduling. Changing demands update the current service loops and thus bring more complexity. Considering changing information in a dynamic manner bring in much more computational complexity. For instance, standard dynamic programming for a dynamic TSP problem with release dates and deadlines has the computational complexity of $O(n^3)$, and more dedicated algorithms still suffer from the $O(n^2)$ complexity~\cite{archetti2015complexity}. Designing more efficient and effective algorithms do bring challenges to the researchs in solving DDS problems via DRL. 
 However, only a limited number of DRL solutions for such a dynamic routing scenario are proposed curretnly~\cite{joe2020deep, shi2019operating， zong2023ai}. Real-time scheduling based on the dynamics is still challenging for DDS development. 

\subsection{AGV and UAV Applications as DDS}
As new tools in both transportation and industry, AGV and UAV are adopted to execute many DDS tasks such as warehousing and food delivery. For example, Hu et al.~\cite{hu2020deep} propose a task dispatching framework to control AGVs for flexible shop floor. However, leveraging totally unmanned service workers brings extra challenges. First, current microscopic control algorithms are not as agile as humans. Even though the dispatching or routing framework generate decisions in a macroscopic way, the unmanned worker need to deal with detailed scenarios during movement. For instance, a human courier can easily avoid a pedestrian without spending much extra time, but a UAV will be slowed down significantly. Such extra expense during inference is difficult to estimate in the training phase, and thus result in performance difference in the simulated environment from real-world scenarios. Second, current UAV and AGV require recharging constantly, modeling the charging actions into dispatching and routing systems becomes a routine, which requires additional designs.
\section{Open Problems}
\label{Sec_opportunities}

With multiple challenges, there are still many open problems with future research opportunities in developing more effective DDS systems. In this section, we briefly discuss some research directions  that we feel may be potentials in this area.

\subsection{Advanced DRL Methods for DDS}
As DRL theory and methods develop rapidly in recent years, new advanced DRL algorithms are of great potential in developing more robust, effective, and efficient DDS applications. For example, leveraging dispatching problems, most literature we investigate in this survey utilizes DQN as the training algorithm. However, such an off-policy framework suffers from limitations to interact with the environment even multi-sourced data is used. Consequently, reproducibility within other environments will face great difficulty and thus cause fairness issues on performance comparison. While recent development on off-line RL~\cite{agarwal2020optimistic} provides new opportunities to tackle the challenges respectively. A complete off-line learning paradigm based on large-scale agent experience data may help to improve training robustness and solve the reproducibility problem. Besides off-line RL, other advanced RL techniques such as causal RL~\cite{gupta2018meta} leveraging multi-objective optimization may also bring new research opportunities into DDS development.

\subsection{Joint Optimization of Two Stages}
Currently, even though both the dispatching stage and the routing stage are well studied, works that consider both stages into a DRL learning paradigm are still missing. As discussed in the previous section, most integrated systems only model one stage using RL and leave the other to a fixed strategy. This forms a major problem, especially in the reaction speed to new changes in applicable systems. For example, current order learning-based dispatching systems are still computationally intensive since conventional VRP solvers are adopted instead of the DRL based~\cite{yu2019integrated, shah2020neural}, which serves as the role to help predict future income and vehicle states. A joint consideration in two stages can help to improve the overall performance, including both planning quality and inference speed. A major challenge of such modeling is the even more complicated state space and the heterogeneity of different action spaces. Research potential lies in the cross-stage representation for both states and actions. The planning quality will be highly related to the hierarchical framework design.

\subsection{Fairness Consideration}
In the current solutions for DDS problems, it almost defaults to set the scheduling objective to maximize the profit of the entire platform. Even though new objectives are proposed, such as Order Response Rate (ORR) in Sec 3.1.1, they still conform to the overall centralized profit. Rather than it, few works stand from the perspective of the service workers. As both DDS and AI ethics develop, researchers of social science gradually focus on how service workers think about their roles in DDS. While keeping optimizing the centralized profit as the prime goal, how to consider the individual differences among the group of service workers and their initiatives is a research question with potentials. On the one hand, fairness between the workers is an essential problem. Maximizing the overall profit alone might result in extreme differences in individual incomes. For instance, \cite{shi2021learning} proposed a ride-hailing method by considering both fairness between different drivers' income along with the overall benefit into their reward design. A well-designed DDS system should guarantee the fairness of different service workers. On the other hand, individual workers might have his or her preference for the dispatching or routing strategies. Personal historical patterns with no intelligent algorithm intervention may help in finding these preferences, and may further be considered as a factor in using intelligent ones to guide them.

\subsection{Partial Compliance Consideration}
In the dispatching stage, current algorithms usually consider full compliance from the service workers as a simplified assumption. However in real-world applications, workers may reject recommendations from the centralized platform and operate based on individual preference. Such disobeying may result in inaccurate overall performance prediction and thus requires additional investigation. 

Besides considering partial compliance as a factor in the system, the reason that causes such compliance and corresponding solutions form another important research task. For example, couriers on rainy days usually have reluctance on accepting distant delivery tasks. A solution is to provide extra allowance to the couriers. Joint decision on generating order matching decisions and determining the specific allowance quota for different couriers based on specific tasks is a sequential decision task, which is suitable for DRL modeling.

\subsection{Large-Scale Online Scheduling System}
Tackling the challenges as discussed in Sec~\ref{Sec_Challengs}, the ultimate benchmark of utilizing DRL in DDS is to build a large-scale online scheduling system to handle real-world DDS tasks. Complete system development requires thorough consideration on solving coupled and dynamic features, modeling heterogeneity within fleets, remaining high efficiency in large scales, and adapting to practical constraints. Both general high-quality and robust algorithms and ad-hoc considerations in specific scenarios are needed in constructing a centralized platform for DDS design. Developing a large-scale online scheduling DDS system via DRL will have a strong impact on both relative research and industrial areas. 
\section{Conclusion}
\label{Sec_conclusion}

Demand driven services (DDS), such as ride-hailing and express systems, are of great importance in urban life nowadays. The planning and scheduling process within these applications require effectiveness and efficiency. In this survey, we focus on the DDS problems and derive the entire DDS into two stages, the dispatching stage and the routing stage respectively. The dispatching stage is responsible to coordinate the unassigned service demands and the available service workers, while the routing stage generates strategies within each service loops. We  investigate the recent works using deep reinforcement learning (DRL) techniques to solve DDS problems in the two stages. We also discuss the further challenges and open problems in using DRL to help build high-quality DDS systems.

\section{Acknowledgment}
This work was supported in part by the National Key Research and Development Program of China under grant 2022ZD0116402, the National Natural Science Foundation of China under 62272260, 62171260.



\bibliographystyle{ACM-Reference-Format}
\bibliography{ref-simple}

\end{document}